\documentclass[runningheads]{llncs}
\usepackage{graphicx}
\usepackage{amsmath,amssymb} 
\usepackage{color}
\usepackage{subcaption}
\usepackage[width=122mm,left=12mm,paperwidth=146mm,height=193mm,top=12mm,paperheight=217mm]{geometry}
\usepackage{comment}
\usepackage[pagebackref=true,breaklinks=true,colorlinks,bookmarks=false]{hyperref}
\usepackage{mathtools}
\usepackage{booktabs}
\usepackage{multirow}

\usepackage{xspace}

\newcommand{\sect}[1]{Section~\ref{#1}}

\newcommand{\vpar}[1]{\paragraph{\normalfont\bf #1}\ \ }

\newcommand{\fig}[1]{Figure~\ref{#1}}
\newcommand{\tbl}[1]{Table~\ref{#1}}
\newcommand{\ignorethis}[1]{}

\newcommand{\eg}{{e.g.}\@\xspace}
\newcommand{\ie}{{i.e.}\@\xspace}
\newcommand{\etal}{{et al.}\@\xspace}

\newcommand{\binmod}{Binary\xspace}
\newcommand{\clustermod}{Clustering\xspace}
\newcommand{\spectmodel}{Spectrum\xspace}

\usepackage{array}
\newcolumntype{L}[1]{>{\raggedright\let\newline\\\arraybackslash\hspace{0pt}}m{#1}}
\newcolumntype{C}[1]{>{\centering\let\newline\\\arraybackslash\hspace{0pt}}m{#1}}
\newcolumntype{R}[1]{>{\raggedleft\let\newline\\\arraybackslash\hspace{0pt}}m{#1}}

\newcommand{\eccvplaceholder}{}

\newcommand{\winsizesec}{3.75\xspace}
\newcommand{\numfullbands}{32\xspace}

\newcommand{\numpervideo}{10\xspace}

\newcommand{\soundtexdim}{502}

\newcommand{\numclusters}{30\xspace}
\newcommand{\testclusteracc}{15.8\%\xspace}
\newcommand{\clusterpurechance}{3.3\%\xspace}
\newcommand{\clustermostcommon}{6.6\%\xspace}
\newcommand{\numtrainimsmil}{1.8\xspace} 

\newcommand{\numobjsununits}{67\xspace}
\newcommand{\numobjtrackingsununits}{61\xspace}

\newcommand{\numobjunits}{91\xspace}
\newcommand{\numplacesunits}{117\xspace}
\newcommand{\numtrackingunits}{72\xspace}
\newcommand{\nummotionunits}{27\xspace}
\newcommand{\fracvidsound}{43.7\%\xspace}

\begin{document}


\newcommand{\fix}{\marginpar{FIX}}
\newcommand{\new}{\marginpar{NEW}}
\newcommand\encoderarrow{\stackrel{\mathclap{\footnotesize \normalfont\mbox{enc}}}{\rightarrow}}
\newcommand\decoderarrow{\stackrel{\mathclap{\footnotesize \normalfont\mbox{dec}}}{\rightarrow}}

\definecolor{MyDarkBlue}{rgb}{0,0.08,1}
\definecolor{MyDarkGreen}{rgb}{0.02,0.6,0.02}
\definecolor{MyDarkRed}{rgb}{0.8,0.02,0.02}
\definecolor{MyDarkOrange}{rgb}{0.40,0.2,0.02}
\definecolor{MyPurple}{RGB}{111,0,255}
\definecolor{MyRed}{rgb}{1.0,0.0,0.0}
\definecolor{MyGold}{rgb}{0.75,0.6,0.12}
\definecolor{MyDarkgray}{rgb}{0.66, 0.66, 0.66}

\pagestyle{headings}
\mainmatter

\title{Ambient Sound Provides Supervision \\for Visual Learning}
\titlerunning{Ambient Sound Provides Supervision for Visual Learning}
\authorrunning{Owens et al.}

\author{Andrew Owens\textsuperscript{1}, Jiajun Wu\textsuperscript{1}, Josh H. McDermott\textsuperscript{1},\\William T. Freeman\textsuperscript{1,2}, and Antonio Torralba\textsuperscript{1}}
\institute{\textsuperscript{1}Massachusetts Institute of Technology\\\textsuperscript{2}Google Research}

\maketitle

\begin{abstract}
The sound of crashing waves, the roar of fast-moving cars~--~sound conveys important information about the objects in our surroundings. In this work, we show that ambient sounds can be used as a supervisory signal for learning visual models.  To demonstrate this, we train a convolutional neural network to predict a statistical summary of the sound associated with a video frame.  We show that, through this process, the network learns a representation that conveys information about objects and scenes.  We evaluate this representation on several recognition tasks, finding that its performance is comparable to that of other state-of-the-art unsupervised learning methods. Finally, we show through visualizations that the network learns units that are selective to objects that are often associated with characteristic sounds. 
\end{abstract}

\section{Introduction}

\begin{figure}[t!]
  \includegraphics[width=\linewidth]{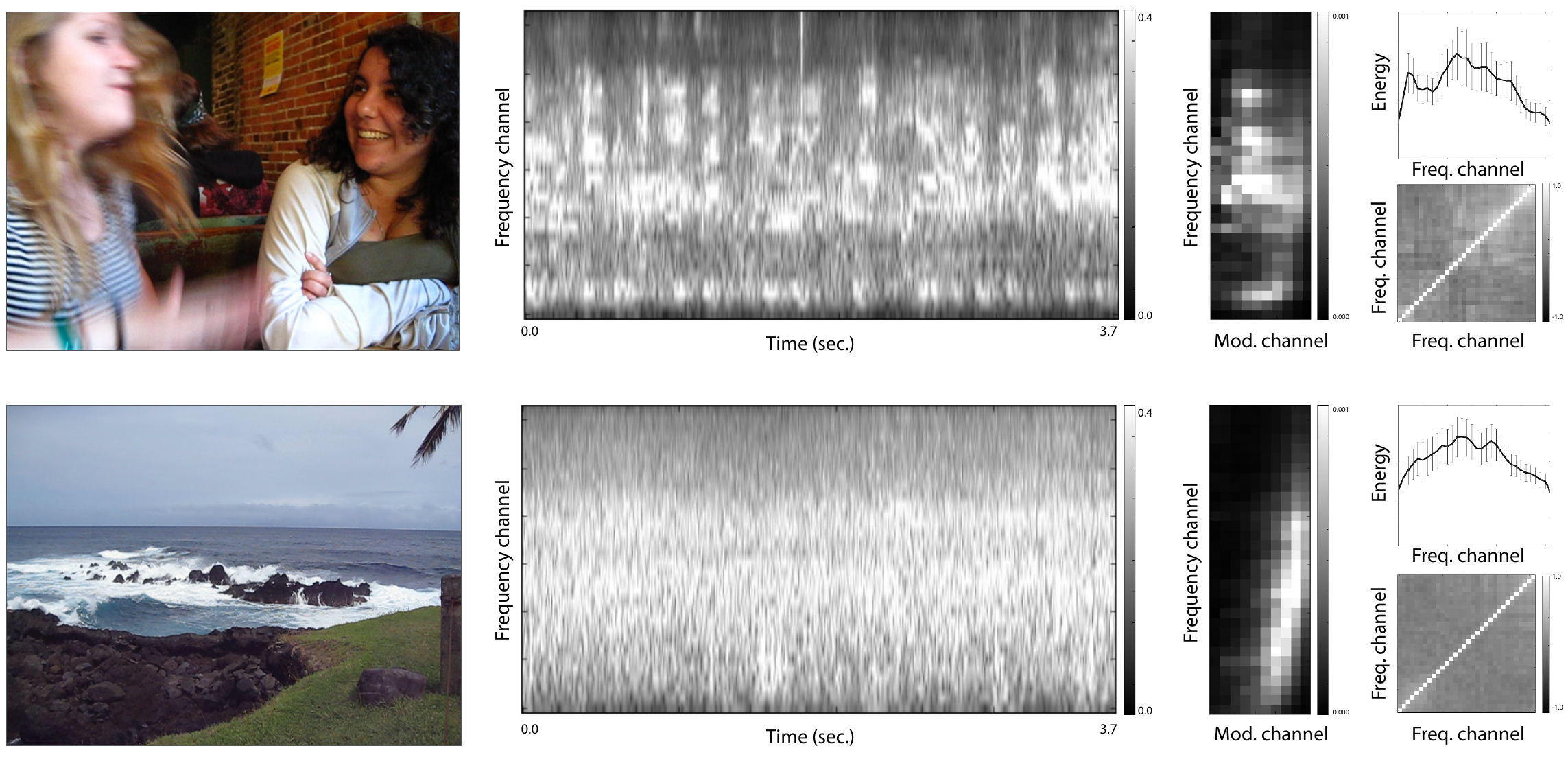}
  {\small \noindent \phantom{~~~~~} (a) Video frame \hspace{19mm} (b) Cochleagram \hspace{14mm} (c) Summary statistics}
  \caption{Visual scenes are associated with characteristic
    sounds. Our goal is to take an image (a) and predict time-averaged
    summary statistics (c) of a cochleagram (b).  The statistics we
    use are (clockwise): the response to a bank of band-pass
    modulation filters (sorted left-to-right in increasing order of
    frequency); the mean and standard deviation of each frequency
    band; and the correlation between bands.  We show two frames from
    the Flickr video dataset \cite{thomee2015yfcc100m}.  The first
    contains the sound of human speech; the second contains the sound
    of wind and crashing waves.  The differences between these sounds
    are reflected in their summary statistics: \eg, the water/wind
    sound, which is similar to white noise, contains fewer
    correlations between cochlear channels. }
  \label{fig:soundtex}
\end{figure}

Sound conveys important information about the world around us -- the bustle of a caf\'{e} tells us that there are many people nearby, while the low-pitched roar of engine noise tells us to watch for fast-moving cars \cite{gaver1993world}. Although sound is in some cases complementary to visual information, such as when we listen to something out of view, vision and hearing are often informative about the same structures in the world. Here we propose that as a consequence of these correlations, concurrent visual and sound information provide a rich training signal that we can use to learn useful representations of the visual world.

In particular, an algorithm trained to predict the sounds that occur within a visual scene might be expected to learn objects and scene elements that are associated with salient and distinctive noises, such as people, cars, and flowing water. Such an algorithm might also learn to associate visual scenes with the ambient sound textures \cite{mcdermott2011sound} that occur within them.  It might, for example, associate the sound of wind with outdoor scenes, and the buzz of refrigerators with indoor scenes.

Although human annotations are indisputably useful for learning, they are expensive to collect. The correspondence between ambient sounds and video is, by contrast, ubiquitous and free.  While there has been much work on learning from unlabeled image data \cite{doersch2015unsupervised,wang2015unsupervised,le2012building}, an audio signal may provide information that that is largely orthogonal to that available in images alone -- information about semantics, events, and mechanics are all readily available from sound \cite{gaver1993world}.

One challenge in utilizing audio-visual input is that the sounds that we hear are only loosely associated with what we see. Sound-producing objects often lie outside of our visual field, and objects that are capable of producing characteristic sounds -- barking dogs, ringing phones -- do not always do so. A priori it is thus not obvious what might be achieved by predicting sound from images.

In this work, we show that a model trained to predict held-out sound from video frames learns a visual representation that conveys semantically meaningful information.  We formulate our sound-prediction task as a classification problem, in which we train a convolutional neural network (CNN) to predict a statistical summary of the sound that occurred at the time a video frame was recorded.  We then validate that the learned representation contains significant information about objects and scenes.

We do this in two ways: first, we show that the image features that we learn through our sound-prediction task can be used for object and scene recognition. On these tasks, our features obtain similar performance to state-of-the-art unsupervised and self-supervised learning methods. Second, we show that the intermediate layers of our CNN are highly selective for objects.  This augments recent work \cite{zhou2014object} showing that object detectors ``emerge'' in a CNN's internal representation when it is trained to recognize scenes.  As in the scene recognition task, object detectors emerge inside of our sound-prediction network. However, our model learns these detectors from an unlabeled audio-visual signal, without any explicit human annotation.

In this paper, we: (1) present a model based on visual CNNs and sound textures \cite{mcdermott2011sound} that predicts a video frame's held-out sound; (2) demonstrate that the CNN learns units in its convolutional layers that are selective for objects, extending the methodology of Zhou \etal \cite{zhou2014object}; (3) validate the effectiveness of sound-based supervision by using the learned representation for object- and scene-recognition tasks.  These results suggest that sound data, which is available in abundance from consumer videos, provides a useful training signal for visual learning.

\section{Related Work}

We take inspiration from work in psychology, such as Gaver's Everyday Listening \cite{gaver1993world}, that studies the ways that humans learn about objects and events using sound.  In this spirit, we would like to study the situations where sound tells us about visual objects and scenes.  Work in auditory scene analysis ~\cite{ellis2011classifying,eronen2006audio,lee2010detecting} meanwhile has provided computational methods for recognizing structures in audio streams.  Following this work, we use a sound representation \cite{mcdermott2011sound} that has been applied to sound recognition \cite{ellis2011classifying} and synthesis tasks \cite{mcdermott2011sound}.

Recently, researchers have proposed many unsupervised learning methods that learn visual representations by solving prediction tasks (sometimes known as {\em pretext} tasks) for which the held-out prediction target is derived from a natural signal in the world, rather than from human annotations. This style of learning has been called ``self supervision'' \cite{doersch2015unsupervised} or ``natural supervision'' \cite{owens2015visually}. With these methods, the supervisory signal may come from video, for example by having the algorithm estimate camera motion \cite{agrawal2015learning,jayaraman2015learning} or track content across frames \cite{wang2015unsupervised,mobahi2009deep,goroshin2015unsupervised}. There are also methods that learn from static images, for example by predicting the relative location of image patches \cite{doersch2015unsupervised,isola2015learning}, or by learning invariance to simple geometric and photometric transformations \cite{dosovitskiy2014discriminative}.  The assumption behind these methods is that, in order to solve the pretext task, the model has to implicitly learn about semantics and, through this process, develop image features that are broadly useful.

While we share with this work the high-level goal of learning image representations, and we use a similar technical approach, our work differs in significant ways.  In contrast to methods whose supervisory signal comes entirely from the imagery itself, ours comes from a modality (sound) that is complementary to vision. This is advantageous because sound is known to be a rich source of information about objects and scenes~\cite{gaver1993world,ellis2011classifying}, and it is largely invariant to visual transformations, such as lighting, scene composition, and viewing angle. Predicting sound from images thus requires some degree of generalization to visual transformations.  Moreover, our supervision task is based on solving a straightforward classification problem, which allows us to use a network design that closely resembles those used in object and scene recognition (rather than, for example, the siamese-style networks used in video methods).

Our approach is closely related to recent audio-visual work \cite{owens2015visually} that predicts soundtracks for videos that show a person striking objects with a drumstick.  A key feature of this work is that the sounds are ``visually indicated'' by actions in video -- a situation that has also been considered in other contexts, such as in the task of visually localizing a sound source \cite{hershey1999audio,kidron2005pixels,fisher2000learning} or in evaluating the synchronization between the two modalities \cite{slaney2000facesync}.  In the natural videos that we use, however, the sound sources are frequently out of frame.  Also, in contrast to other recent work in multi-modal representation learning~\cite{ngiam2011multimodal,srivastava2012multimodal,andrew2013deep}, our technical approach is based on solving a self-supervised classification problem (rather than a generative model or autoencoder), and our goal is to learn visual representations that are generally useful for object recognition tasks.


\begin{figure}[t!]



\eccvplaceholder
\includegraphics[width=\linewidth]{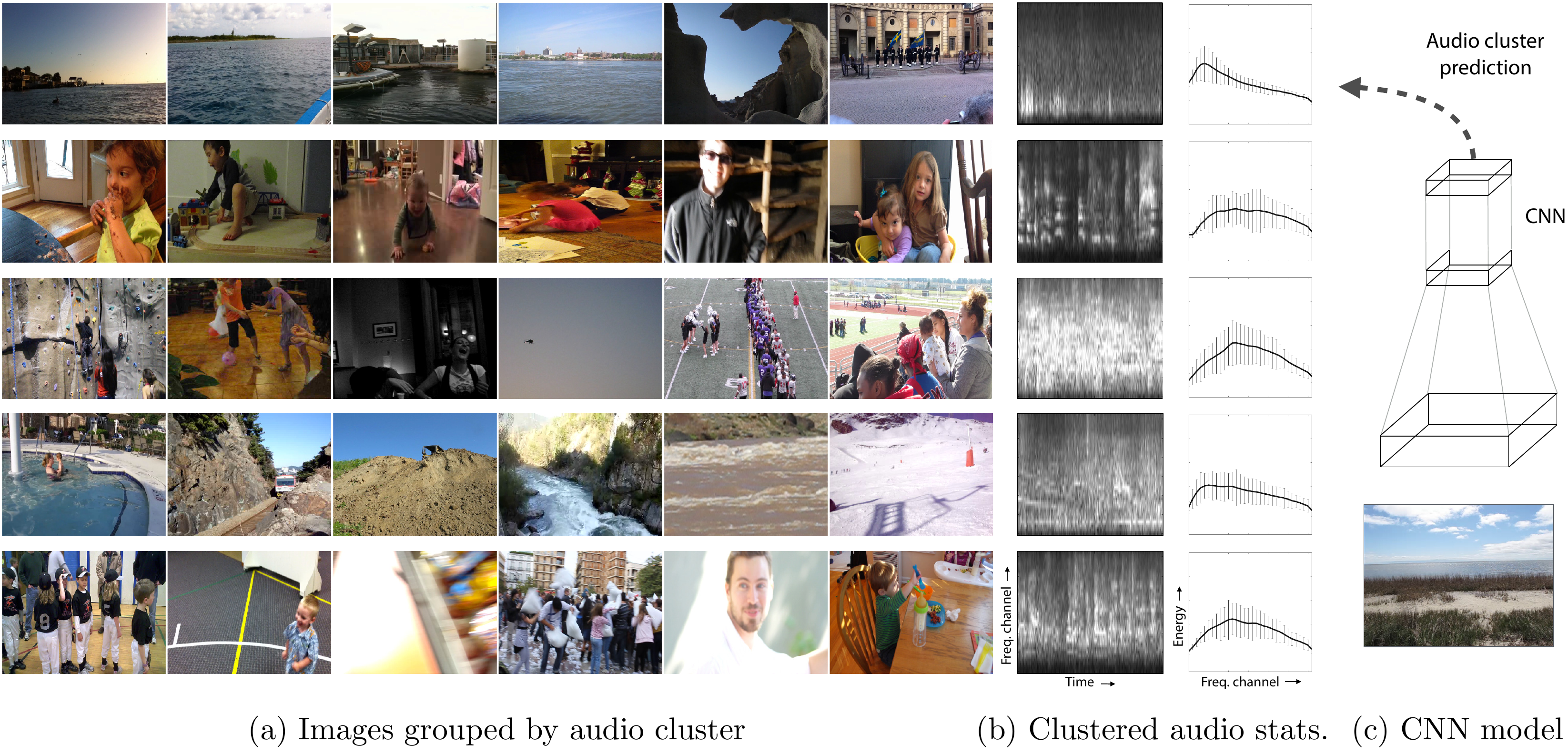}
\caption{Visualization of some of the audio clusters used in one of our models (5 of 30 clusters).  For each cluster, we show (a) the images in the test set whose sound textures were closest to the centroid (no more than one frame per video), and (b) we visualize aspects of the sound texture used to define the cluster centroid -- specifically, the mean and standard deviation of the frequency channels.  We also include a representative cochleagram  (that of the leftmost image).  Although the clusters were defined using audio, there are common objects and scene attributes in many of the images.  We train a CNN to predict a video frame's auditory cluster assignment (c).}
\label{fig:soundclusters}
  
\end{figure}

\section{Learning to predict ambient audio}

\label{sec:predict}

We would like to train a model that, when given a frame of video, can predict its corresponding sound -- a task that implicitly requires knowledge of objects and scenes.

\subsection{Statistical sound summaries} A natural question, then, is how our model should represent sound. Perhaps the first approach that comes to mind would be to estimate a frequency spectrum at the moment in which the picture was taken, similar to \cite{owens2015visually}.  However, this is potentially suboptimal because in natural scenes it is difficult to predict the precise timing of a sound from visual information.  Upon seeing a crowd of people, for instance, we might expect to hear the sound of speech, but the precise timing and content of that speech might not be directly indicated by the video frames.  

To be closer to the time scale of visual objects, we estimate a statistical summary of the sound, averaged over a few seconds.  We do this using the sound texture model of McDermott and Simoncelli~\cite{mcdermott2011sound}, which assumes that sound is stationary within a temporal window (we use \winsizesec seconds).  More specifically, we closely follow~\cite{mcdermott2011sound} and filter the audio waveform with a bank of \numfullbands band-pass filters intended to mimic human cochlear frequency selectivity. We then take the Hilbert envelope of each channel, raise each sample of the envelope to the 0.3 power (to mimic cochlear amplitude compression), and resample the compressed envelope to 400 Hz. Finally, we compute time-averaged statistics of these subband envelopes: we compute the mean and standard deviation of each frequency channel, the mean squared response of each of a bank of modulation filters applied to each channel, and the Pearson correlation between pairs of channels. For the modulation filters, we use a bank of 10 band-pass filters with center frequencies ranging from 0.5 to 200 Hz, equally spaced on a logarithmic scale. 

To make the sound features more invariant to gain (e.g., from the microphone), we divide the envelopes by the median energy (median vector norm) over all timesteps, and include this energy as a feature. As in~\cite{mcdermott2011sound}, we normalize the standard deviation of each cochlear channel by its mean, and each modulation power by its standard deviation.  We then rescale each kind of texture feature (\ie marginal moments, correlations, modulation power, energy) inversely with the number of dimensions. The sound texture for each image is a \soundtexdim-dimensional vector. In \fig{fig:soundtex}, we give examples of these summary statistics for two audio clips.  We provide more details about our audio representation in the supplementary material.


\subsection{Predicting sound from images}

We would like to predict sound textures from images -- a task that we hypothesize leads to learning useful visual representations.  Although multiple frames are available, we predict sound from a single frame, so that the learned image features will be more likely to transfer to single-image recognition tasks.  Furthermore, since the the actions that produce the sounds may not appear on-screen, motion information may not always be applicable.

While one option would be to regress the sound texture $v_j$ directly from the corresponding image $I_j$, we choose instead to define explicit sound categories and formulate this visual recognition problem as a classification task. This also makes it easier to analyze the network, because it allows us to compare the internal representation of our model to object- and scene-classification models with similar network architecture (\sect{sec:objdet}). We consider two labeling models: one based on a vector quantization, the other based on a binary coding scheme.


\vpar{Clustering audio features} In the {\em \clustermod} model, the sound textures $\{v_j\}$ in the training set are clustered using $k$-means.  These clusters define image categories: we label each sound texture with the index of the closest centroid, and train our CNN to label images with their corresponding labels.  

 
We found that audio clips that belong to a cluster often contain common objects.  In \fig{fig:soundclusters}, we show examples of such clusters, and in the supplementary material we provide their corresponding audio.  We can see that there is a cluster that contains indoor scenes with children in them -- these are relatively quiet scenes punctuated with speech sounds. Another cluster contains the sounds of many people speaking at once (often large crowds); another contains many water scenes (usually containing loud wind sounds).  Several clusters capture general scene attributes, such as outdoor scenes with light wind sounds.  During training, we remove examples that are far from the centroid of their cluster (more than the median distance to the vector, amongst all examples in the dataset).

\vpar{Binary coding model} For the other variation of our model (which we call the {\em \binmod} model), we use a binary coding scheme \cite{indyk1998approximate,salakhutdinov2009semantic,weiss2009spectral} equivalent to a multi-label classification problem.  We project each sound texture $v_j$ onto the top principal components (we use 30 projections), and convert these projections into a binary code by thresholding them.  We predict this binary code using a sigmoid layer, and during training we measure error using cross-entropy loss.

For comparison, we trained a model (which we call the {\em \spectmodel} model) to approximately predict the frequency spectrum at the time that the photo was taken, in lieu of a full sound texture.  Specifically, for our sound vectors $v_j$ in this model, we used the mean value of each cochlear channel within a 33.3-millisecond interval centered on the input frame (approximately one frame of a 30 Hz video).  For training, we used the projection scheme from the \binmod model.

\vpar{Training} We trained our models to predict audio on a 360,000-video subset of the Flickr video dataset~\cite{thomee2015yfcc100m}.  Most of the videos in the dataset are personal video recordings containing natural audio, though many were post-processed, \eg with added subtitles, title screens, and music.  We divided our videos into training and test sets, and we randomly sampled \numpervideo frames per video (\numtrainimsmil million training images total). For our network architecture, we used the CaffeNet architecture \cite{jia2014caffe} (a variation of Krizhevsky \etal \cite{krizhevsky2012imagenet}) with batch normalization \cite{ioffe2015batch}.  We trained our model with Caffe \cite{jia2014caffe}, using a batch size of 256, for 320,000 iterations of stochastic gradient descent.


\begin{figure}[t!]
    \centering
  \eccvplaceholder
  \vspace{0.5mm}Training by sound (\numobjunits Detectors) \\
  \vspace{-0.4mm}
  \includegraphics[width=\linewidth]{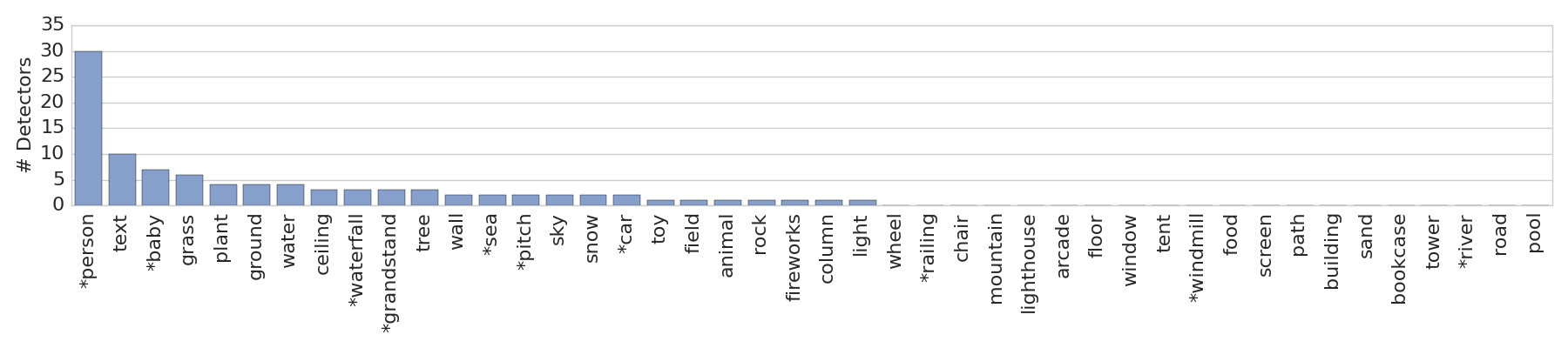}
  Training by labeled scenes~\cite{zhou2014places} (\numplacesunits Detectors) \\
  \vspace{-0.4mm}
  \includegraphics[width=\linewidth]{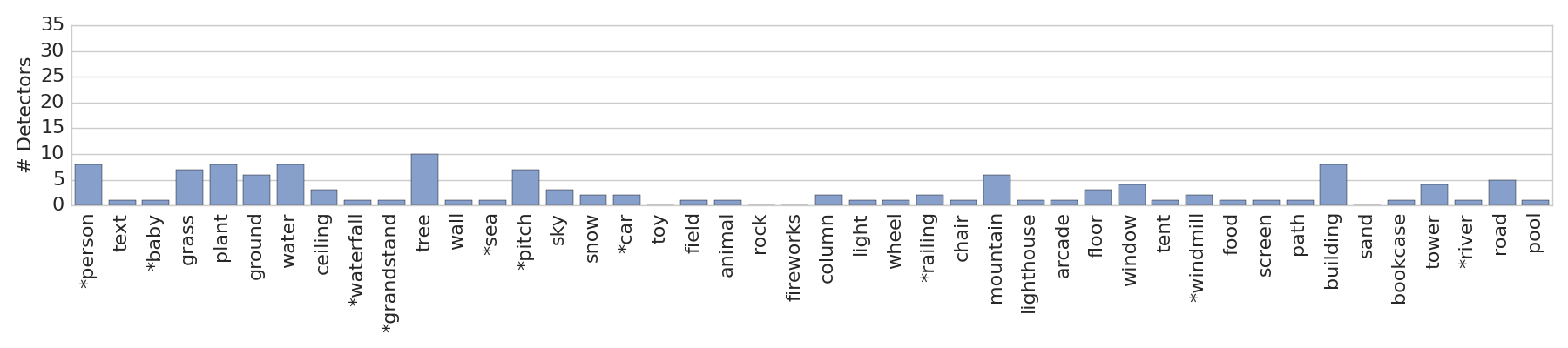}
  Training by visual tracking~\cite{wang2015unsupervised} (\numtrackingunits Detectors)\\
  \vspace{-0.4mm}
  \includegraphics[width=\linewidth]{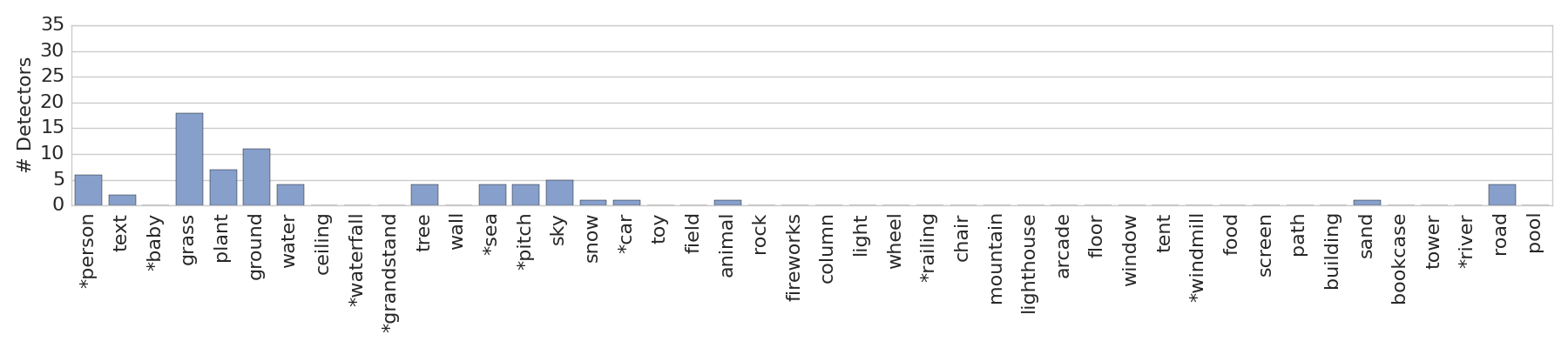}
  \caption{Histogram of object-selective units in networks trained with different styles of supervision. From top to bottom: training to predict ambient sound (our Clustering model); training to predict scene category using the Places dataset~\cite{zhou2014places}; and training to do visual tracking~\cite{wang2015unsupervised}. Compared to the tracking model, which was also trained without semantic labels, our network learns more high-level object detectors. It also has more detectors for objects that make characteristic sounds, such as {\em person}, {\em baby}, and {\em waterfall}, in comparison to the one trained on Places~\cite{zhou2014places}. Categories marked with $*$ are those that we consider to make characteristic sounds.}
  \label{fig:objdistr}
\end{figure}

\begin{figure}[t!]
\scriptsize
\begin{tabular}{C{0.325\textwidth}C{0.325\textwidth}C{0.325\textwidth}}
  \eccvplaceholder
     \multicolumn{3}{c}{Neuron visualizations of the network trained by \bf sound} \\
     field & sky & grass \\
     \includegraphics[width=0.97\linewidth]{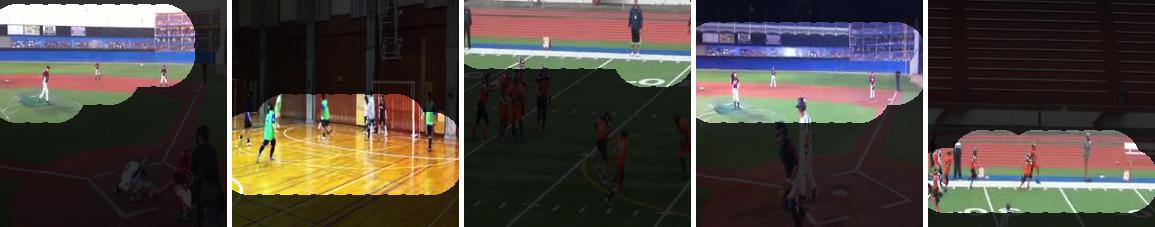} 
     & \includegraphics[width=0.97\linewidth]{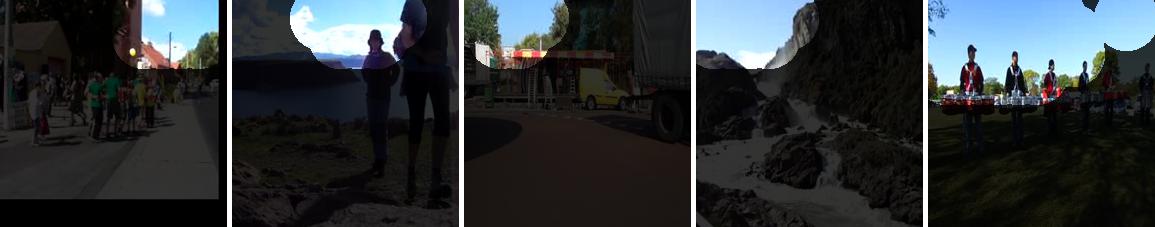} 
     & \includegraphics[width=0.97\linewidth]{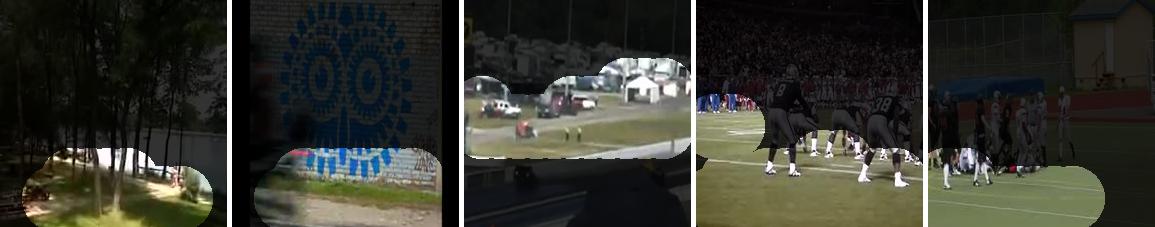} \\
     snowy ground & ceiling & car\\
     \includegraphics[width=0.97\linewidth]{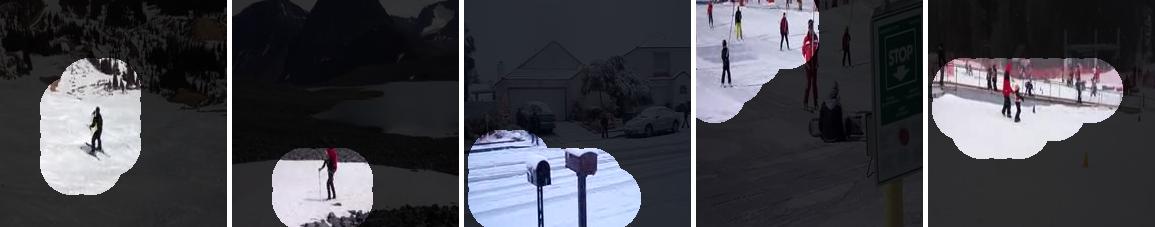}
     & \includegraphics[width=0.97\linewidth]{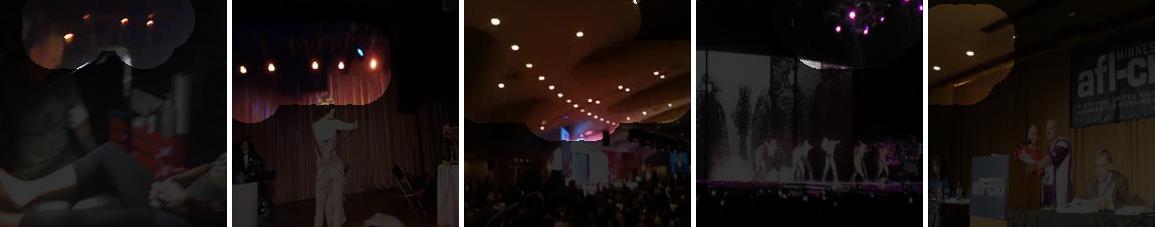} 
     & \includegraphics[width=0.97\linewidth]{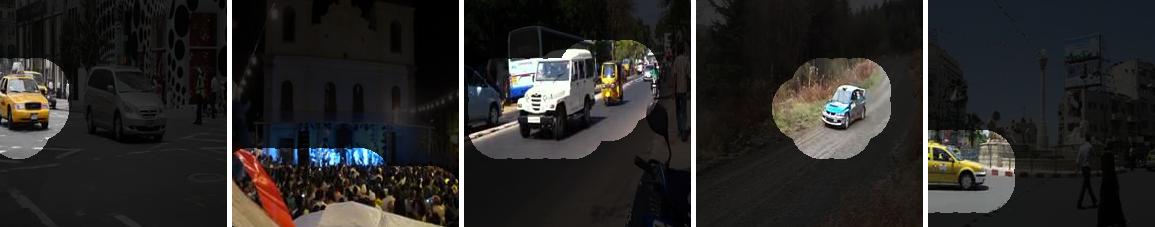} \\
     waterfall & waterfall & sea \\
     \includegraphics[width=0.97\linewidth]{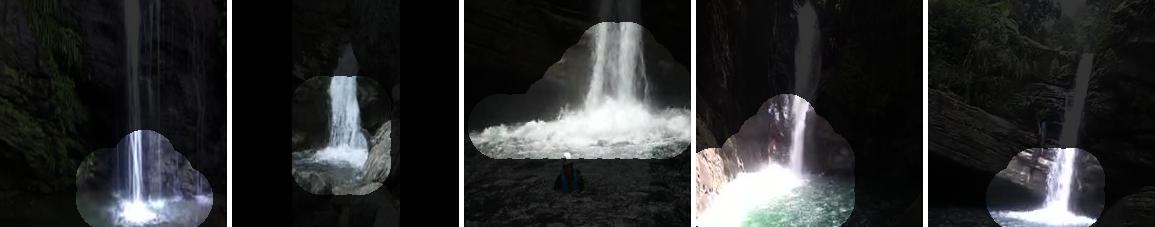}
     & \includegraphics[width=0.97\linewidth]{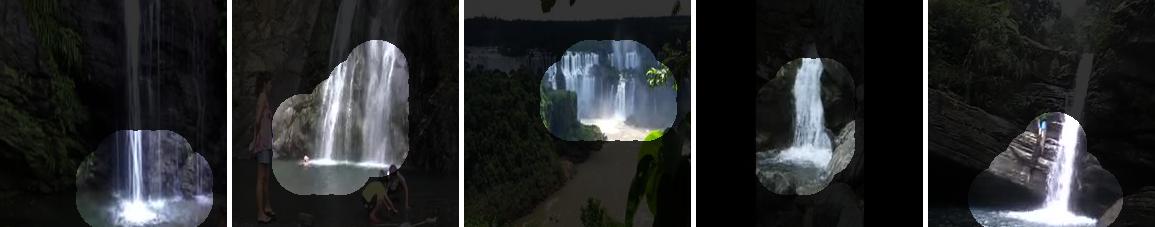}
     & \includegraphics[width=0.97\linewidth]{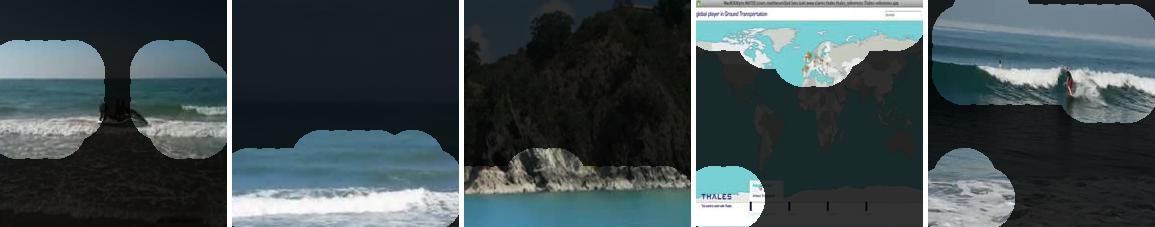} \\
      baby &  baby &  baby\\
     \includegraphics[width=0.97\linewidth]{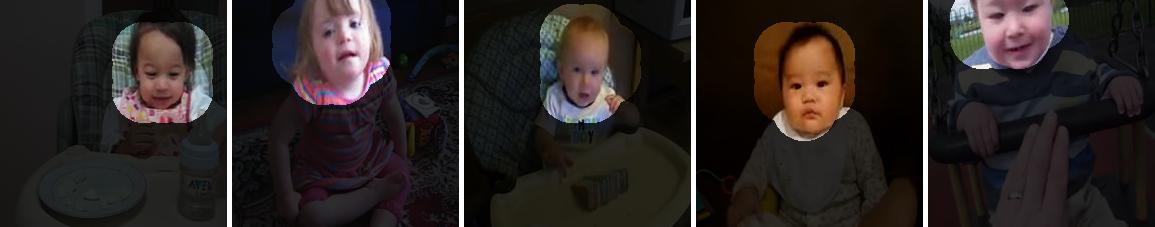}
     & \includegraphics[width=0.97\linewidth]{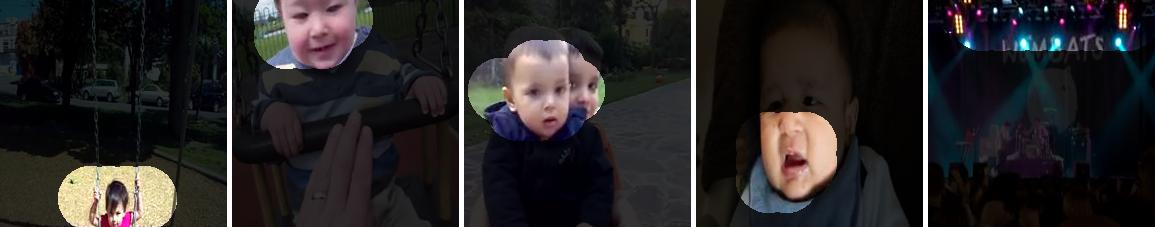}
     & \includegraphics[width=0.97\linewidth]{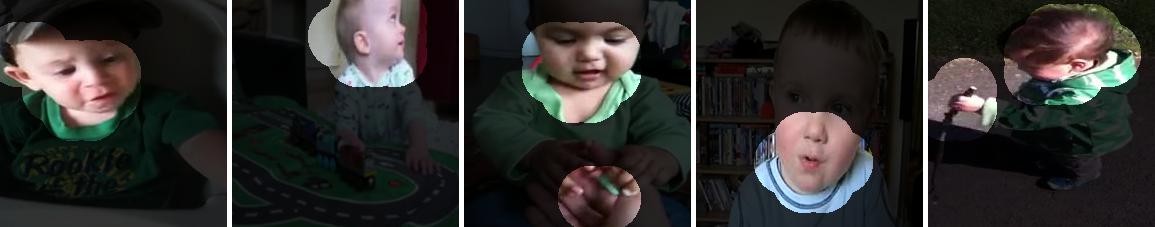} \\
      person &  person &  person \\
     \includegraphics[width=0.97\linewidth]{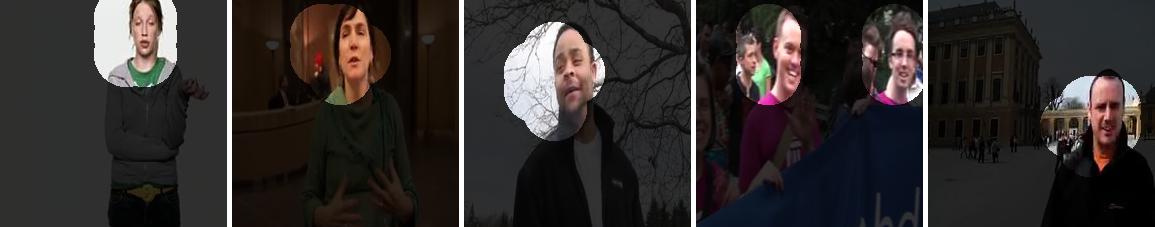}
     & \includegraphics[width=0.97\linewidth]{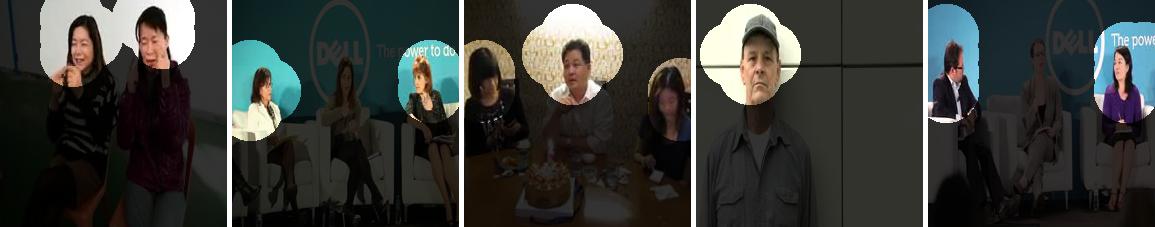}
     & \includegraphics[width=0.97\linewidth]{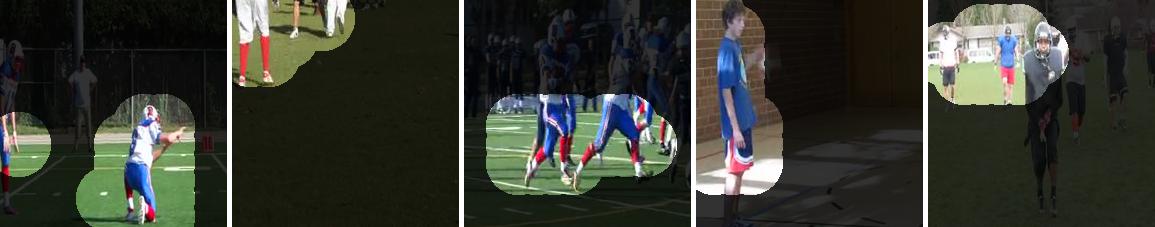} \\
      person &  person &  person  \\
     \includegraphics[width=0.97\linewidth]{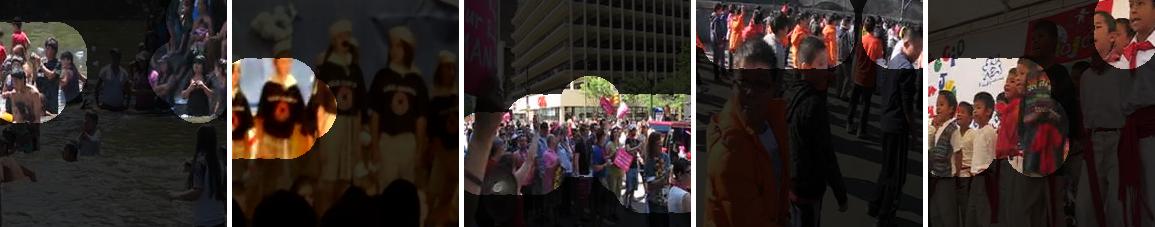}
     & \includegraphics[width=0.97\linewidth]{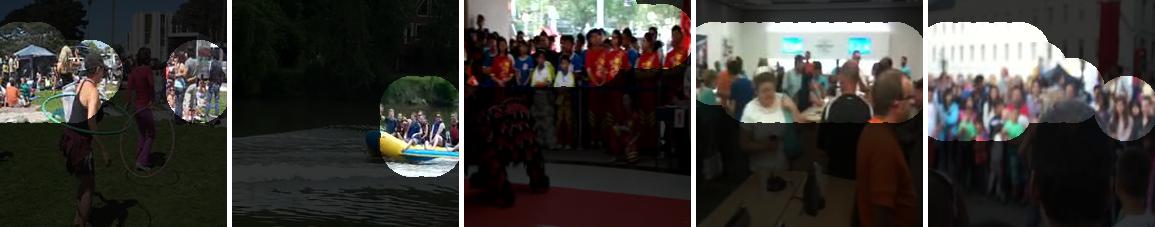}
     & \includegraphics[width=0.97\linewidth]{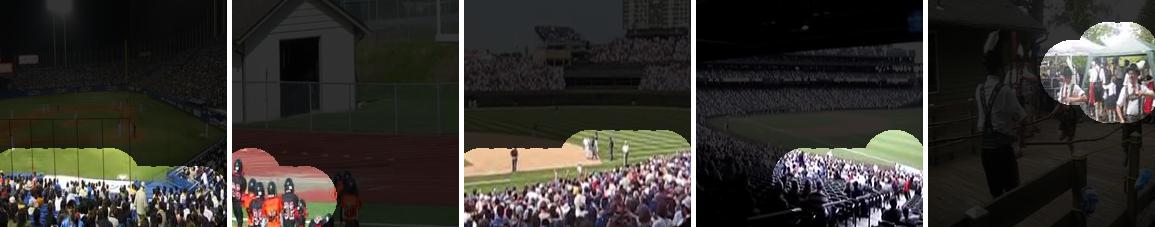} \\
      grandstand &  grandstand &  grandstand \\
     \includegraphics[width=0.97\linewidth]{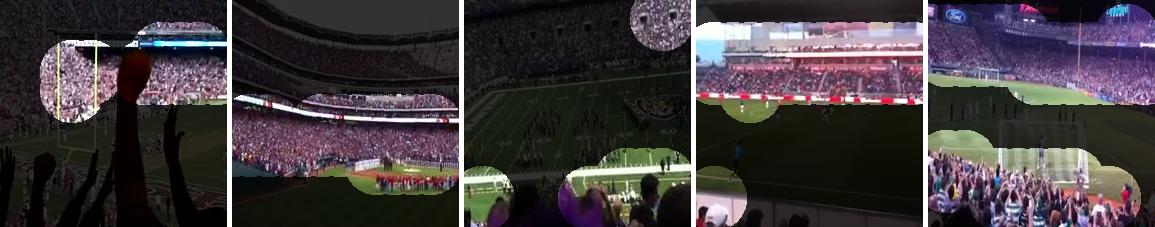}
     & \includegraphics[width=0.97\linewidth]{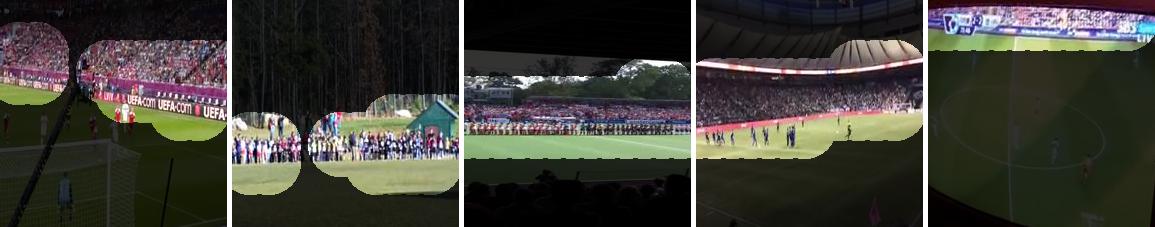}
     & \includegraphics[width=0.97\linewidth]{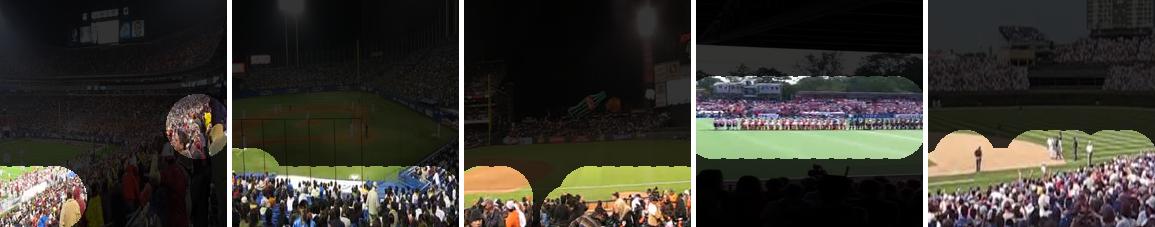} \\
     \midrule
     \multicolumn{3}{c}{Neuron visualizations of the network trained by visual \bf tracking~\cite{wang2015unsupervised}} \\
      sea &  grass &  road \\
     \includegraphics[width=0.97\linewidth]{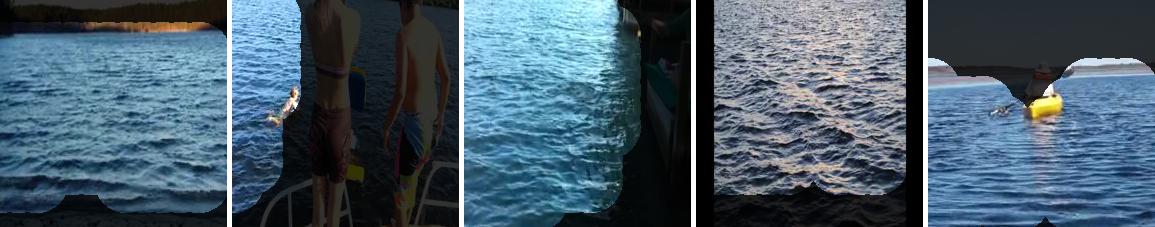} 
     & \includegraphics[width=0.97\linewidth]{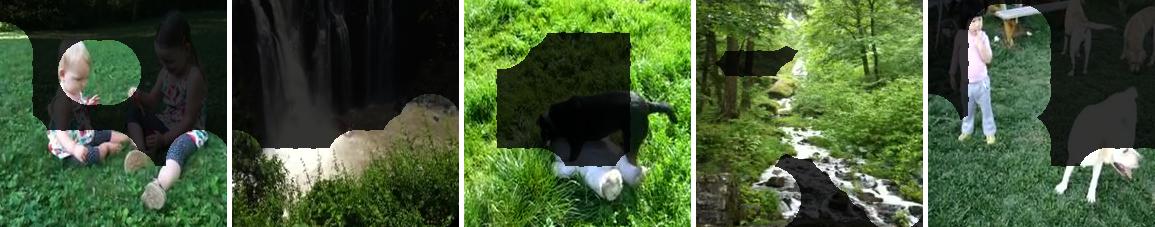} 
     & \includegraphics[width=0.97\linewidth]{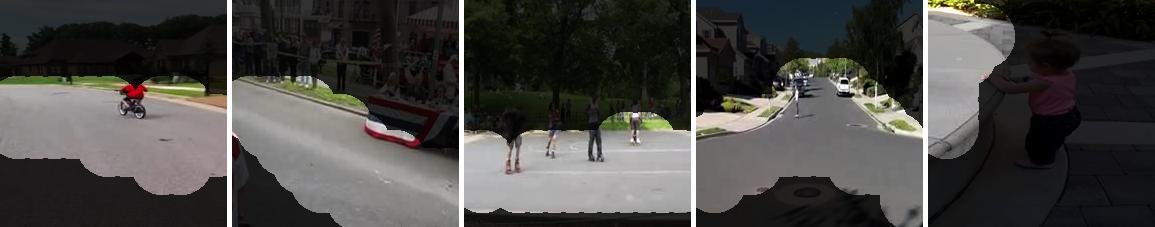} \\
      sea &  pitch &  sky \\
     \includegraphics[width=0.97\linewidth]{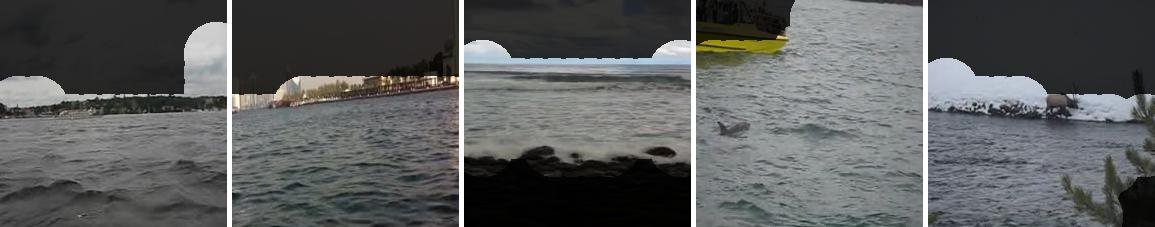}
     & \includegraphics[width=0.97\linewidth]{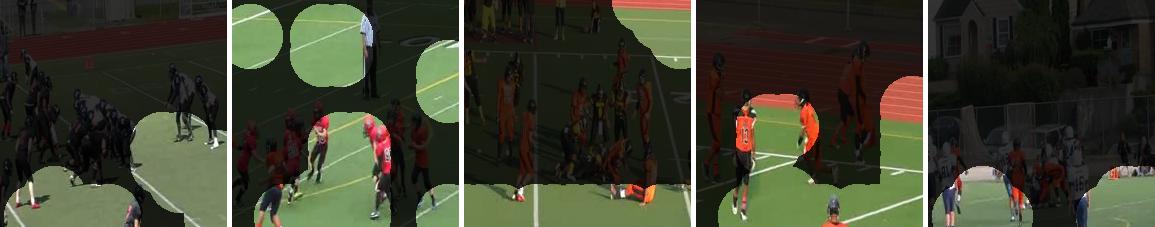}
     & \includegraphics[width=0.97\linewidth]{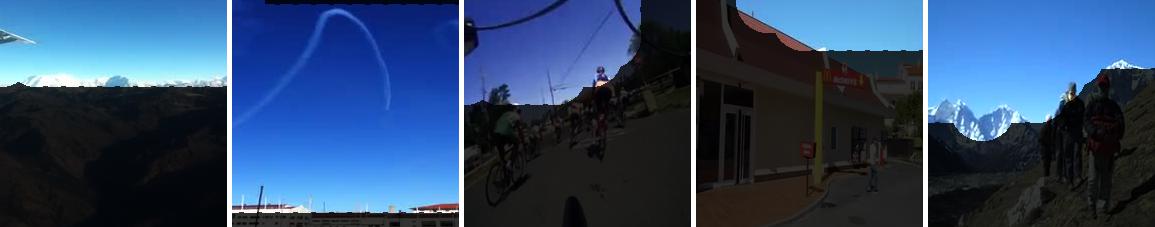} \\
     \midrule
     \multicolumn{3}{c}{Neuron visualizations of the network trained by \bf egomotion~\cite{agrawal2015learning}} \\
      ground &  sky &  grass\\
     \includegraphics[width=0.97\linewidth]{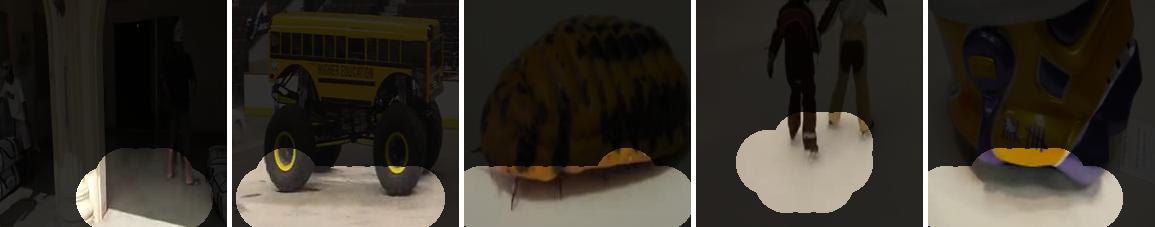} 
     & \includegraphics[width=0.97\linewidth]{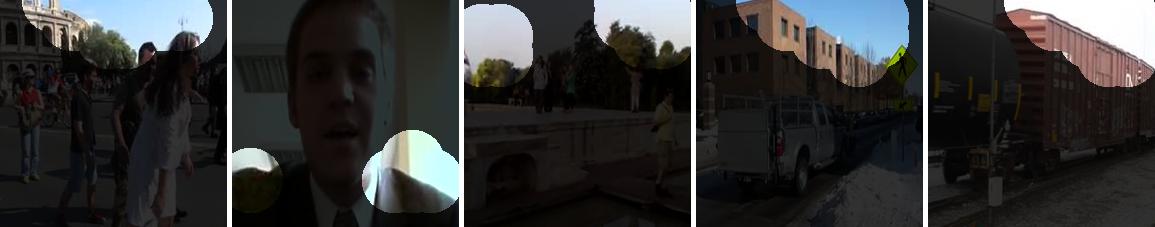} 
     & \includegraphics[width=0.97\linewidth]{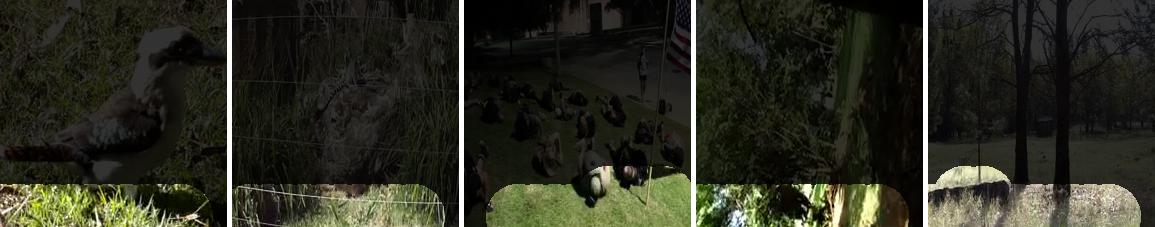} \\
      ground &  sky &  plant\\
     \includegraphics[width=0.97\linewidth]{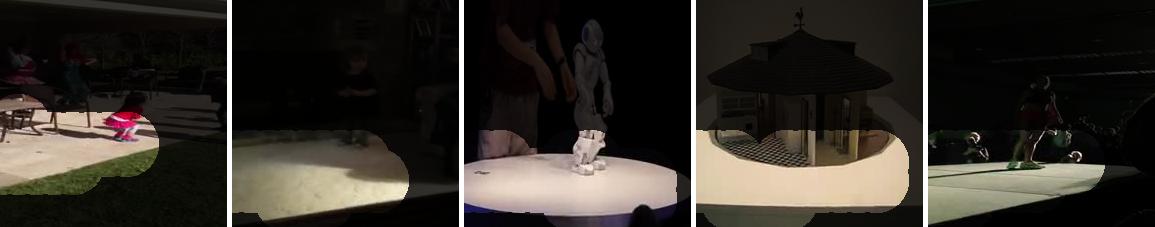}
     & \includegraphics[width=0.97\linewidth]{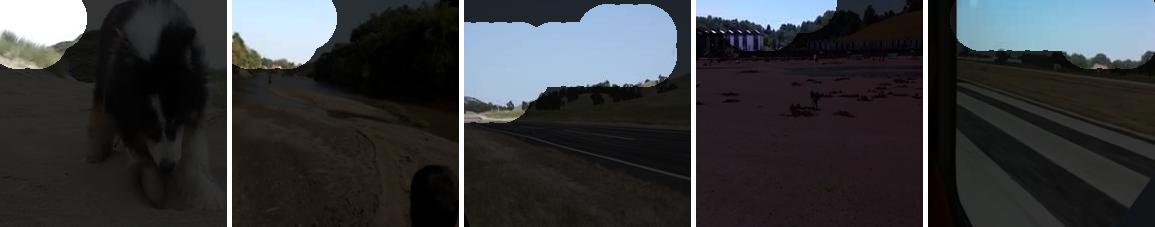}
     & \includegraphics[width=0.97\linewidth]{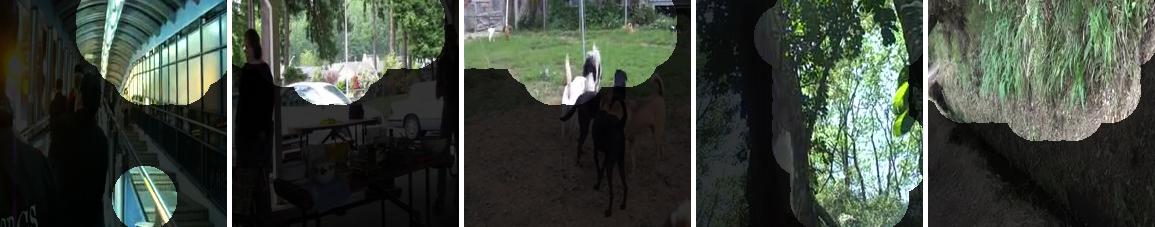} \\
     \midrule
     \multicolumn{3}{c}{Neuron visualizations of the network trained by \bf patch positions~\cite{doersch2015unsupervised}} \\
      sky &  sky &  baby\\
     \includegraphics[width=0.97\linewidth]{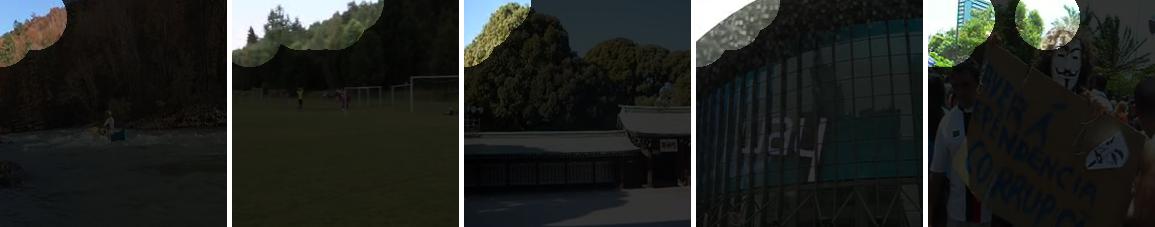}
     & \includegraphics[width=0.97\linewidth]{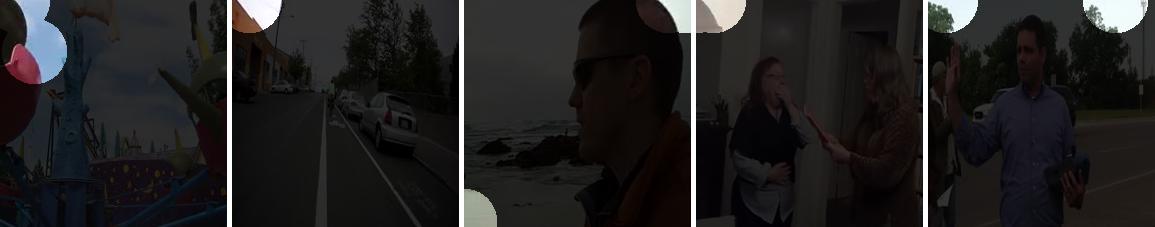}
     & \includegraphics[width=0.97\linewidth]{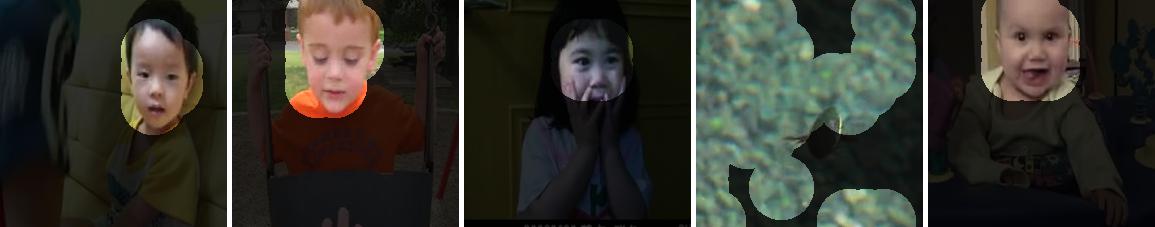} \\
     \midrule
     \multicolumn{3}{c}{Neuron visualizations of the network trained by labeled \bf scenes~\cite{zhou2014places}} \\
      field &  tent &  building\\
     \includegraphics[width=0.97\linewidth]{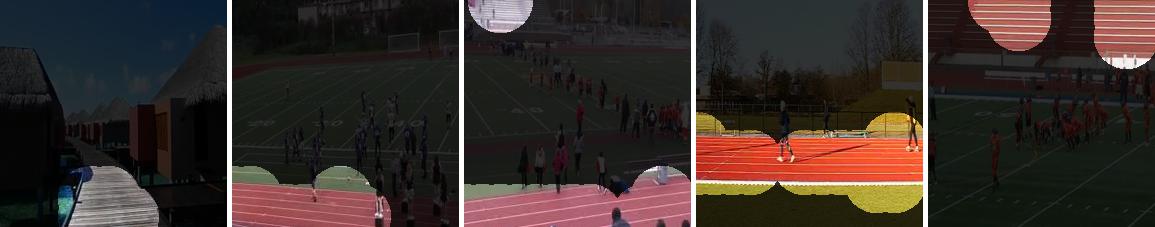} 
     & \includegraphics[width=0.97\linewidth]{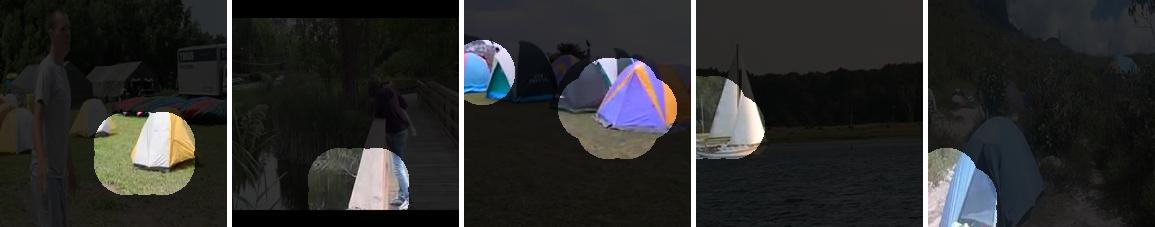} 
     & \includegraphics[width=0.97\linewidth]{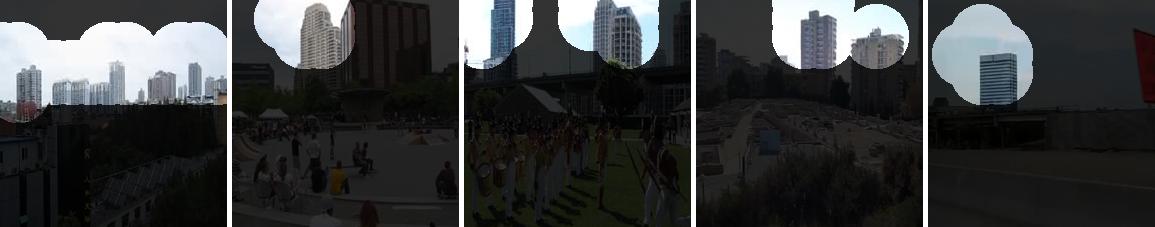} \\
      pitch & path &  sky\\
     \includegraphics[width=0.97\linewidth]{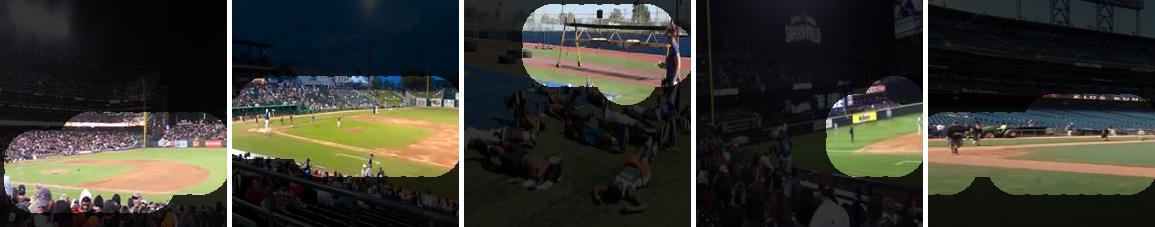} 
     & \includegraphics[width=0.97\linewidth]{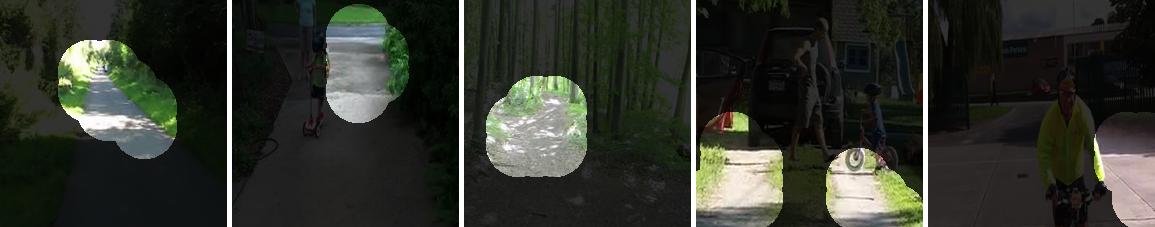} 
     & \includegraphics[width=0.97\linewidth]{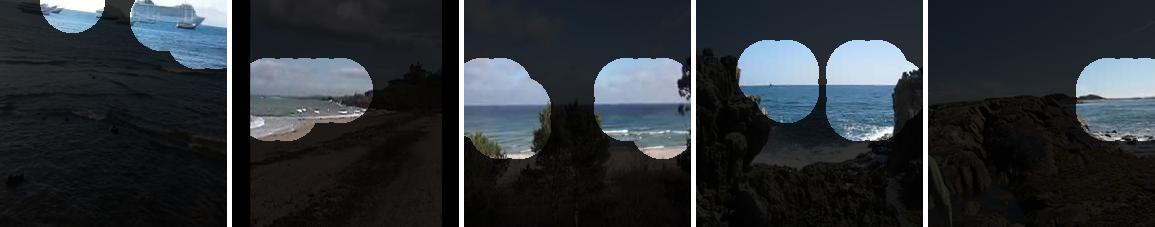}
\end{tabular}
\vspace{-5pt}
  \caption{\small Top 5 responses for neurons of various networks, tested on the Flickr dataset. 
  Please see \sect{sec:extravis} for more visualizations.}
  \label{fig:neuronsamples}
  \vspace{-20pt}
\end{figure}

\section{Results}
\label{sec:results}

We evaluate the image representation that our model learned in multiple ways. First, we demonstrate that the internal representation of our model contains convolutional units (neurons) that are selective to particular objects, and we analyze those objects' distribution.  We then empirically evaluate the quality of the learned representation for several image recognition tasks, finding that it achieves performance comparable to other feature-learning methods that were trained without human annotations.

\subsection{What does the network learn to detect?}
\label{sec:objdet}Previous work~\cite{zhou2014object} has shown that a CNN trained to predict scene categories will learn convolutional units that are selective for objects -- a result that follows naturally from the fact that scenes are often defined by the objects that compose them.  We ask whether a model trained to predict ambient sound, rather than explicit human labels, would learn object-selective units as well. For these experiments, we used our Clustering model, because its network structure is similar to that of the scene-recognition model used in \cite{zhou2014object}.

\vpar{Quantifying object-selective units} Similar to the method in~\cite{zhou2014object}, we visualized the images that each neuron in the top convolutional layer (conv5) responded most strongly to.  To do this, we sampled a pool of 200,000 images from our Flickr video test set.  We then collected, for each convolutional unit, the 60 images in this set that gave the unit the largest activation. Next, we applied the so-called synthetic visualization technique of \cite{zhou2014object} to approximately superimpose the unit's receptive field onto the image.  Specifically, we found all of the spatial locations in the layer for which the unit's activation strength was at least half that of its maximum response.  We then masked out the parts of the image that were not covered by the receptive field of one of these high-responding spatial units. We assumed a circle-shaped receptive field, obtaining the radius from \cite{zhou2014object}.  To examine the effect of the data used in the evaluation, we also applied this visualization technique to other datasets (please see the supplementary material).

Next, for each neuron we showed its masked images to three human annotators on Amazon Mechanical Turk, and asked them: (1) whether an object is present in many of these regions, and if so, what it is; (2) to mark the images whose activations contain these objects. Unlike \cite{zhou2014object}, we only considered units that were selective to objects, ignoring units that were selective to textures.  For each unit, if at least $60\%$ of its top 60 activations contained the object, we considered it to be selective for the object (or following \cite{zhou2014object}, we say that it is a {\em detector} for that object).  We then manually labeled the unit with an object category, using the category names provided by the SUN database \cite{xiao2010sun}.  We found that \numobjunits of the 256 units in our model were object-selective in this way, and we show a selection of them in~\fig{fig:neuronsamples}.

We compared the number of these units to those of a CNN trained to recognize human-labeled scene categories on Places~\cite{zhou2014object}.  As expected, this model -- having been trained with explicit human annotations -- contained more object-selective units (\numplacesunits units).  We also asked whether object-selective neurons appear in the convolutional layers when a CNN is trained on other tasks that do not use human labels. As a simple comparison, we applied the same methodology to the egomotion-based model of Agrawal \etal~\cite{agrawal2015learning} and to the tracking-based method of Wang and Gupta~\cite{wang2015unsupervised}.  We applied these networks to whole images (in all cases resizing the input image to $256 \times 256$ pixels and taking the center $227 \times 227$ crop), though we note that they were originally trained on cropped image regions.

We found that the tracking-based method also learned object-selective units, but that the objects that it detected were often textural ``stuff,'' such as grass, ground, and water, and that there were fewer of these detection units in total (\numtrackingunits of $256$).  The results were similar for the egomotion-based model, which had \nummotionunits such units.  In \fig{fig:objdistr} and in the supplementary material, we provide the distribution of the objects that the units were selective to.  We also visualized neurons from the method of Doersch \etal~\cite{doersch2015unsupervised} (as before, applying the network to whole images, rather than to patches).  We found a significant number of the units were selective for position, rather than to objects.  For example, one convolutional unit responded most highly to the upper-left corner of an image -- a unit that may be useful for the training task, which involves predicting the relative position of image patches.  In \fig{fig:neuronsamples}, we show visualizations of a selection of object-detecting neurons for all of these methods.

The differences between the objects detected by these methods and our own may have to do with the requirements of the tasks being solved.  The other unsupervised methods, for example, all involve comparing multiple input images or sub-images in a relatively fine-grained way. This may correspondingly change the representation that the network learns in its last convolutional layer -- requiring its the units to encode, say, color and geometric transformations rather than object identities.  Moreover, these networks may represent semantic information in other (more distributed) ways that would not necessarily be revealed through this visualization method.


\begin{table}[t!]
    \centering
  \eccvplaceholder
  \begin{tabular}{lC{1.5cm}C{1.5cm}}
  \toprule
  Method & Sound & Places \\
  \midrule
       \# Detectors & \numobjunits & \numplacesunits \\
       \# Detectors for objects with characteristic sounds & 49 & 26\\
       Videos with object sound & \fracvidsound & 16.9\% \\
       Characteristic sound rate & 81.2\% & 75.9\%\\
    \bottomrule
  \end{tabular}
  \vspace{5pt}
  \caption{Row 1: the number of detectors (\ie units that are selective to a particular object); row 2: the number of detectors for objects with characteristic sounds; row 3: fraction of videos in which an object's sound is audible (computed only for object classes with characteristic sounds); row 4: given that an activation corresponds to an object with a characteristic sound, the probability that its sound is audible. There are 256 units in total for each method.}
  \label{tbl:objsound}
\end{table}

\vpar{Analyzing the types of objects that were detected} 
Next, we asked what kinds of objects our network learned to detect.  We hypothesized that the object-selective neurons were more likely to respond to objects that produce (or are closely associated with) characteristic sounds. To evaluate this, we (an author) labeled the SUN object categories according to whether they were closely associated with a characteristic sound. We denote these categories with a $*$ in \fig{fig:objdistr}. Next, we counted the number of units that were selective to these objects, finding that our model contained significantly more such units than a scene-recognition network trained on the Places dataset, both in total number and as a proportion (\tbl{tbl:objsound}). A significant fraction of these units were selective to people (adults, babies, and crowds). 

Finally, we asked whether the sounds that these objects make were actually present in the videos that these video frames were sampled from.  To do this, we listened to the sound of the top 30 video clips for each unit, and recorded whether the sound was made by the object that the neuron was selective to (\eg, human speech for the {\em person} category).  We found that \fracvidsound of these videos contained the objects' sounds (\tbl{tbl:objsound}).

\subsection{Evaluating the image representation}
\label{sec:feateval}

\begin{table}[t!]
  \centering
  \scriptsize
  \begin{subtable}[b]{.69\linewidth}
  \begin{tabular}{l C{.08\linewidth}C{.08\linewidth}C{.08\linewidth}C{.08\linewidth} C{.08\linewidth}C{.08\linewidth}C{.08\linewidth}C{.08\linewidth} }
        \toprule
        \multirow{2}{*}{Method} & \multicolumn{4}{c}{VOC Cls. (\%mAP)} & \multicolumn{4}{c}{SUN397 (\%acc.)} \\
        \cmidrule(r{2pt}){2-5}  \cmidrule(l{2pt}){6-9}
        & {max5} & {pool5} & {fc6} & {fc7}    & {max5} & {pool5} & {fc6} & {fc7}\\
        \midrule
        Sound (cluster)                                & 36.7 & 45.8 & 44.8 & 44.3                           & {\bf 17.3} & {\bf 22.9} & 20.7 & 14.9   \\
        Sound (binary)                                 & {\bf 39.4} & {\bf 46.7} & {\bf 47.1} & {\bf 47.4}   & 17.1 & 22.5 & {\bf 21.3} & {\bf 21.4}         \\
        Sound (spect.)                                 & 35.8 & 44.0 & 44.4 & 44.4                           & 14.6 & 19.5 & 18.6 & 17.7      \\
        Texton-CNN                                     & 28.9 & 37.5 & 35.3 & 32.5                           & 10.7 & 15.2 & 11.4 & 7.6      \\
        K-means \cite{krahenbuhl2015data}              & 27.5 & 34.8 & 33.9 & 32.1                           & 11.6 & 14.9 & 12.8 & 12.4      \\
        Tracking~\cite{wang2015unsupervised}           & 33.5 & 42.2 & 42.4 & 40.2                           & 14.1 & 18.7 & 16.2 & 15.1      \\
        Patch pos.~\cite{doersch2015unsupervised}      & 27.7 & {\bf 46.7} & - & -                           & 10.0 & 22.4 & - & -     \\
        Egomotion~\cite{agrawal2015learning}           & 22.7 & 31.1 & - & -                                 & 9.1  & 11.3 & - & -     \\
        \midrule                                        
        ImageNet~\cite{krizhevsky2012imagenet}         & {\bf 63.6} & {\bf 65.6} & {\bf 69.6} & {\bf 73.6}   & 29.8 & 34.0 & 37.8 & 37.8   \\
        Places~\cite{zhou2014places}                   & 59.0 & 63.2 & 65.3 & 66.2                           & {\bf 39.4} & {\bf 42.1} & {\bf 46.1} & {\bf 48.8}    \\
        \bottomrule
  \end{tabular}
  \caption{\label{fig:classify} Image classification with linear SVM}
  \end{subtable} %
  \begin{subtable}[b]{.3\linewidth}
  \scriptsize
      \begin{tabular}{lc}
      \toprule
      Method  & (\%mAP) \\
      \midrule
      Random init. \cite{krahenbuhl2015data} & 41.3 \\
      Sound (cluster)  & 44.1 \\ 
      Sound (binary)  & 43.3 \\ 
      Motion~\cite{wang2015unsupervised,krahenbuhl2015data} & 47.4 \\
      Egomotion~\cite{agrawal2015learning,krahenbuhl2015data} & 41.8 \\
      Patch pos.~\cite{doersch2015unsupervised,krahenbuhl2015data} & 46.6 \\
      Calib. + Patch  \cite{doersch2015unsupervised,krahenbuhl2015data} & {\bf 51.1} \\
      \midrule
      ImageNet~\cite{krizhevsky2012imagenet} & {\bf 57.1} \\ 
      Places~\cite{zhou2014places} & 52.8 \\ 
      \bottomrule
  \end{tabular}
  \caption{\label{fig:detect} Finetuning detection}
  \end{subtable}
  \begin{subtable}[b]{\linewidth}
    \scriptsize\centering
 
    \begin{tabular}{lcccccccccccccccccccc}
\toprule
Method & aer & bk~ & brd & bt~ & btl & bus & car & cat & chr & cow & din & dog & hrs & mbk & prs & pot & shp & sfa & trn & tv~ \\
\midrule

Sound (cluster) & 68 & {\bf 47} & 38 & 54 & 15 & 45 & {\bf 66} & 45 & 42 & 23 & 37 & 28 & 73 & 58 & {\bf 85} & 25 & 26 & 32 & 67 & 42\\
Sound (binary) & 69 & 45 & 38 & 56 & {\bf 16} & {\bf 47} & 65 & 45 & 41 & {\bf 25} & 37 & 28 & {\bf 74} & {\bf 61} & {\bf 85} & 26 & {\bf 39} & 32 & {\bf 69} & 38\\
Sound (spect.) & 65 & 40 & 35 & 54 & 14 & 42 & 63 & 41 & 39 & 24 & 32 & 25 & 72 & 56 & 81 & {\bf 27} & 33 & 28 & 65 & 40\\
Texton-CNN & 65 & 35 & 28 & 46 & 11 & 31 & 63 & 30 & 41 & 17 & 28 & 23 & 64 & 51 & 74 & 9 & 19 & 33 & 54 & 30\\
K-means & 61 & 31 & 27 & 49 & 9 & 27 & 58 & 34 & 36 & 12 & 25 & 21 & 64 & 38 & 70 & 18 & 14 & 25 & 51 & 25\\
Motion~\cite{wang2015unsupervised} & 67 & 35 & 41 & 54 & 11 & 35 & 62 & 35 & 39 & 21 & 30 & 26 & 70 & 53 & 78 & 22 & 32 & 37 & 61 & 34\\
Patches~\cite{doersch2015unsupervised} & {\bf 70} & 44 & {\bf 43} & {\bf 60} & 12 & 44 & {\bf 66} & {\bf 52} & {\bf 44} & 24 & {\bf 45} & {\bf 31} & 73 & 48 & 78 & 14 & 28 & {\bf 39} & 62 & {\bf 43}\\
Egomotion~\cite{agrawal2015learning} & 60 & 24 & 21 & 35 & 10 & 19 & 57 & 24 & 27 & 11 & 22 & 18 & 61 & 40 & 69 & 13 & 12 & 24 & 48 & 28\\

\midrule
ImageNet~\cite{krizhevsky2012imagenet} & 79 & {\bf 71} & {\bf 73} & 75 & {\bf 25} & 60 & 80 & {\bf 75} & 51 & {\bf 45} & 60 & {\bf 70} & {\bf 80} & {\bf 72} & {\bf 91} & 42 & {\bf 62} & 56 & 82 & 62\\
Places~\cite{zhou2014places} & {\bf 83} & 60 & 56 & {\bf 80} & 23 & {\bf 66} & {\bf 84} & 54 & {\bf 57} & 40 & {\bf 74} & 41 & {\bf 80} & 68 & 90 & {\bf 50} & 45 & {\bf 61} & {\bf 88} & {\bf 63}\\
\bottomrule
\end{tabular}
\caption{\label{fig:classify-class}Per class mAP for image classification on PASCAL VOC 2007}
\end{subtable}
  \caption{(a) Mean average precision for PASCAL VOC 2007 classification, and accuracy on SUN397. Here we trained a linear SVM using the
    top layers of different networks. We note in \sect{sec:feateval} that the shape of these layers varies between networks.  (b) Mean average precision on PASCAL VOC 2007 using Fast-RCNN \cite{girshick2015fast}. We initialized the CNN weights using those of our learned sound models.  (c) Per-class AP scores for the VOC 2007 classification task with pool5 features (corresponds to mAP in (a)).}
\end{table}

We have seen through visualizations that a CNN trained to predict sound from an image learns units that are highly selective for objects.  Now we evaluate, experimentally, how well the CNN's internal representation conveys information that is useful for recognizing objects and scenes.

Since our goal is to measure the amount of semantic information provided by the learned representation, rather than to seek absolute performance, we used a simple evaluation scheme.  In most experiments, we computed image features using our CNN and trained a linear SVM to predict object or scene category using the activations in the top layers.

\vpar{Object recognition} First, we used our CNN features for object
recognition on the PASCAL VOC 2007 dataset
\cite{everingham2010pascal}.  We trained a one-vs.-rest linear SVM to
detect the presence of each of the 20 object categories in the dataset, using the activations of the upper layers of the network as the feature set (pool5, fc6, and fc7). To help understand whether the convolutional units considered in \sect{sec:objdet} directly convey semantics, we also created a global max-pooling feature (similar to \cite{oquab2015object}), where we applied max pooling over the entire convolutional layer.  This produces a 256-dimensional vector that contains the maximum response of each convolutional unit (we call it {\em max5}). Following common practice, we evaluated the network on a center $227\times227$ crop of each image (after resizing the image to $256\times256$), and we evaluated the results using mean average precision (mAP).  We chose the SVM regularization parameter for each method by maximizing mAP on the validation set using grid search (we used $\{0.5^k \mid 4 \leq k < 20\}$).

The other unsupervised (or self-supervised) models in our comparison \cite{doersch2015unsupervised,agrawal2015learning,wang2015unsupervised} use different network designs.  In particular, \cite{doersch2015unsupervised} was trained on image patches, so following their experiments we resized its convolutional layers for $227\times227$ images and removed the model's fully connected layers\footnote{As a result, this model has a larger pool5 layer than the other methods: $7 \times 7$ vs. $6 \times 6$. Likewise, the fc6 layer of \cite{wang2015unsupervised} is smaller (1,024 dims. vs. 4,096 dims.).}.  Also, since the model of Agrawal
\etal \cite{agrawal2015learning} did not have a pool5 layer, we added
one to it.  We also considered CNNs that were trained with human
annotations: object recognition on ImageNet \cite{deng2009imagenet}
and scene categories on Places
\cite{zhou2014places}.  Finally, we considered using the $k$-means weight initialization method of \cite{krahenbuhl2015data} to set the weights of a CNN model (we call this the {\em K-means} model).

We found that our best-performing of our model (the binary-coding method) obtained comparable performance to other unsupervised learning methods, such as \cite{doersch2015unsupervised}\footnote{Since the initial version, we have updated the performance numbers for the method of Doersch \etal \cite{doersch2015unsupervised} in \tbl{fig:classify} and Wang \etal \cite{wang2015unsupervised} in \tbl{fig:detect}.  Please see the paper updates in \sect{sec:discussion} for details.} Both models based on sound textures (Clustering and Binary) outperformed the model that predicted only the frequency spectrum.  This suggests that the additional time-averaged statistics from sound textures are helpful.  For these models, we used \numclusters clusters (or PCA projections): in \fig{fig:cluster-curve}, we consider varying the number of clusters, finding that there is a small improvement from increasing it, and a substantial decrease in performance when using just two clusters.  The sound-based models significantly outperformed other methods when we globally pooled the conv5 features, suggesting that the convolutional units contain a significant amount of semantic information (and are well suited to being used at this spatial scale).

\vpar{Scene recognition} We also evaluated our model on a scene
recognition task using the SUN dataset \cite{xiao2010sun}, a large classification benchmark that involves recognizing 397 scene categories with 7,940 training and test images provided in multiple splits.  Following~\cite{agrawal2015learning}, we averaged our classification accuracy across 3 splits, with 20 examples per scene category.  We chose the linear SVM's regularization parameter for each model using 3-fold cross-validation.
    
We again found that our features' performance was comparable to other models.  In particular, we found that the difference between our models was smaller than in the object-recognition case, with both the Clustering and Binary models obtaining performance comparable to the patch-based method with pool5 features.  

\vpar{Pretraining for object detection} Following recent work
\cite{wang2015unsupervised,doersch2015unsupervised,krahenbuhl2015data},
we used our model to initialize the weights of a CNN-based object detection system (Fast R-CNN \cite{girshick2015fast}), verifying that the results improved over random initialization.  We followed the training procedure of Kr\"ahenb\"uhl \etal \cite{krahenbuhl2015data}, using 150,000 iterations of backpropagation with an initial learning rate of 0.002, and we compared our model with other published results (we report the numbers provided by \cite{krahenbuhl2015data}).

Our best-performing model (the Clustering model) obtains similar performance to
that of Wang and Gupta's tracking-based model
\cite{wang2015unsupervised}, while the overall best results were from variations of Doersch \etal's patch-based model \cite{doersch2015unsupervised,krahenbuhl2015data}.  We note that the network changes substantially during fine-tuning, and thus the performance is fairly dependent on the parameters used in the training procedure. Moreover all models, when fine-tuned in this way, achieve results that are close to those of a well-chosen random initialization (within 6\% mAP).  Recent work
\cite{krahenbuhl2015data,mishkin2015all} has addressed these optimization issues by
rescaling the weights of a pretrained network using a data-driven procedure. The unsupervised method with the best performance combines the rescaling method of \cite{krahenbuhl2015data} with the patch-based pretraining of \cite{doersch2015unsupervised}.

\vpar{Sound prediction} We also asked how well our model learned to
solve its sound prediction task.  We found that on our test set, the clustering-based model (with \numclusters clusters) chose the correct sound label \testclusteracc of the time.  Pure chance in this case is
\clusterpurechance, while the baseline of choosing the most commonly
occurring label is \clustermostcommon.  

\vpar{Audio supervision} It is natural to ask what role audio plays in the learning process.  Perhaps, for example, our training procedure would produce equally good features if we replaced the hand-crafted sound features with hand-crafted {\em visual} features, computed from the images themselves.  To study this, we replaced our sound texture features with (512-dimensional) visual texton histograms \cite{leung2001representing}, using the parameters from \cite{xiao2010sun}, and we used them to train a variation of our Clustering model.  

As expected, the images that belong to each cluster are visually coherent, and share common objects.  However, we found that the network performed significantly worse than the audio-based method on the object- and scene-recognition metrics (\tbl{fig:classify}).  Moreover, we found that its convolutional units rarely were selective for objects (generally they responded responded to ``stuff'' such as grass and water). Likely this is because the network simply learned to approximate the texton features, obtaining low labeling error without high-level generalization.  In contrast, the audio-based labels -- despite also being based on another form of hand-crafted feature -- are largely invariant to visual transformations, such as lighting and scale, and therefore predicting them requires some degree of generalization (one benefit of training with multiple, complementary modalities).

\section{Discussion}
\label{sec:discussion}
Sound has many properties that make it useful as a supervisory
training signal: it is abundantly available without human annotations,
and it is known to convey information about objects and scenes.  It is
also complementary to visual information, and may
therefore convey information not easily obtainable from unlabeled
image analysis.

In this work, we proposed using ambient sound to learn visual
representations.  We introduced a model, based on convolutional neural
networks, that predicts a statistical sound summary from a video
frame.  We then showed, with visualizations and experiments on
recognition tasks, that the resulting image representation contains
information about objects and scenes.

Here we considered one audio representation, based on sound textures,
but it is natural to ask whether other audio representations would
lead the model to learn about additional types of objects.  To help
answer this question, we would like to more systematically study the
situations when sound does (and does not) tell us about objects in the
visual world.  Ultimately, we would like to know what object and scene
structures are detectable through sound-based training, and we see our
work as a step in this direction.

\vpar{Acknowledgments} This work was supported by NSF grants \#1524817 to A.T;  NSF grants \#1447476 and \#1212849 to W.F.; a McDonnell Scholar Award to J.H.M.; and a Microsoft
Ph.D. Fellowship to A.O.  It was also supported by Shell Research, and
by a donation of GPUs from NVIDIA. We thank Phillip Isola for the
helpful discussions, and Carl Vondrick for sharing the data that we
used in our experiments. We also thank the anonymous reviewers for
their comments (in particular, for suggesting the comparison with
texton features in \sect{sec:feateval}).

\vspace{6mm}
\noindent\textbf{Paper updates:}\\

\noindent {\bf v1}~ ECCV camera-ready version.

\vspace{2.5mm}
\noindent {\bf v2} In \fig{fig:soundtex}, we now sort the modulation channels by increasing
frequency (thanks to Dan Ellis for pointing this out).  We fixed an
error in the comparison with
Doersch \etal \cite{doersch2015unsupervised} in \tbl{fig:classify} (we
used it with the wrong ordering of the color channels in the initial
version).  We updated the detection accuracy (\tbl{fig:detect})
for \cite{wang2015unsupervised}, which we obtained from the updated
version of \cite{krahenbuhl2015data}. We also fixed a misspelling
in \fig{fig:objdistr}.

\bibliographystyle{splncs03}
\bibliography{ambient}
\renewcommand{\thesection}{A\arabic{section}}
\renewcommand{\thefigure}{A\arabic{figure}}
\setcounter{section}{0}
\setcounter{figure}{0}

\section{Sound label space}
Why do detectors for certain objects -- e.g., people, water, and infants -- emerge in our model? To help answer this question, we visualized the audio clusters that are used to define our model's label space (\sect{sec:predict}). The results, including sound clips, are provided on our webpage.  In \fig{fig:cluster-curve}, we also examine how the quality of the learned image features varies as a function of the number of clusters, as measured by performance on the object recognition task.

\section{Additional unit visualizations}
\label{sec:extravis}
\begin{figure}[t!]
  \centering
  {\scriptsize
  \begin{subfigure}[b]{0.495\linewidth}
    \hspace{4mm}\includegraphics[width=0.84\linewidth]{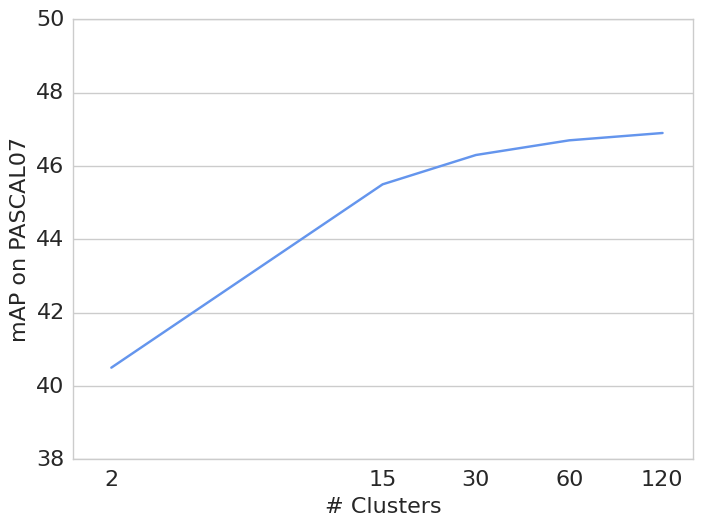} 
    \caption{Varying the number of clusters} \label{fig:cluster-curve}
  \end{subfigure}
  \begin{subfigure}[b]{0.495\linewidth}
    \includegraphics[width=0.98\linewidth]{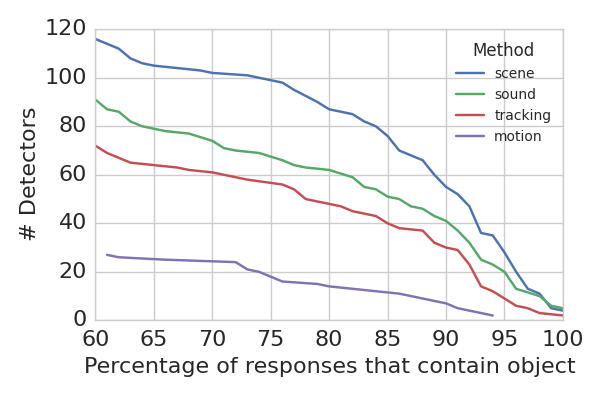}
    \caption{Varying the threshold for selectivity} \label{fig:threshold-curve}
  \end{subfigure}
  }
\caption{(a) Object recognition performance (recognition performance
  on PASCAL VOC2007) increases with the number of clusters used to
  define the audio label space.  For our experiments, we used 30
  clusters. (b) The number of object-selective units for each method,
  as we increase the threshold used to determine whether a unit is
  object-selective.  This threshold corresponds to the fraction of
  images that contain the object in question, amongst the images with
  the 60 largest activations.  For our analysis in \sect{sec:results},
  we used a threshold of 60\%.}
\end{figure}

\begin{figure}[t!]
    \centering
  Training by egomotion~\cite{agrawal2015learning} (\nummotionunits detectors) \\
  \includegraphics[width=\linewidth]{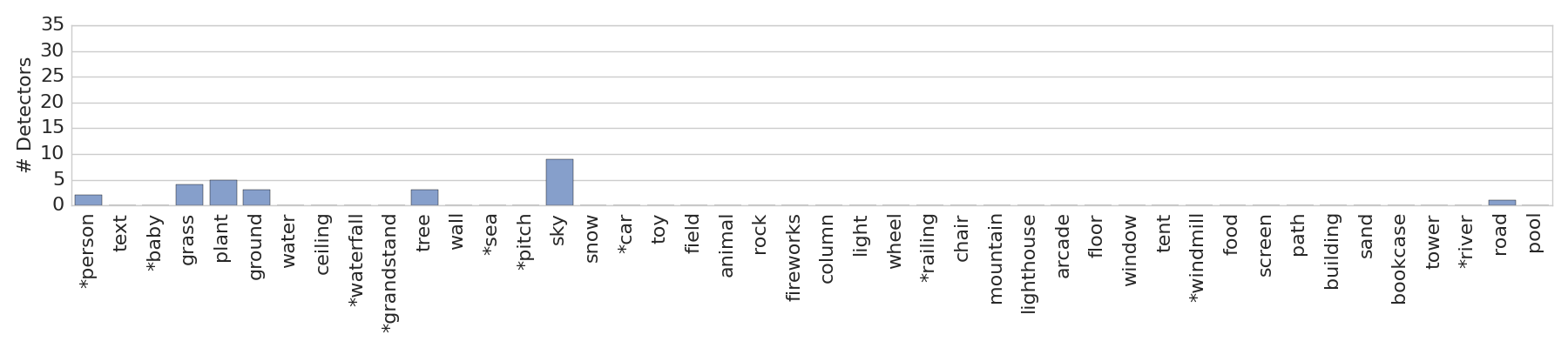}\\
  \caption{Here we quantify the number of object-selective units for an additional method, using the Flickr video dataset (cf. \fig{fig:objdistr}).}
  \label{fig:objdistrsupp}
\end{figure}

\begin{figure}[t!]
\scriptsize\centering
\begin{tabular}{ccc}
person&person&person\\
\includegraphics[width=0.325\linewidth]{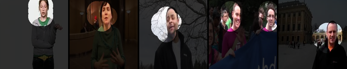}
&\includegraphics[width=0.325\linewidth]{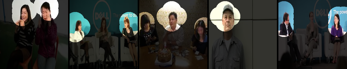}
&\includegraphics[width=0.325\linewidth]{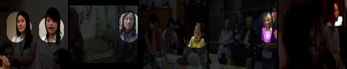}
\\
person&person
&person\\
\includegraphics[width=0.325\linewidth]{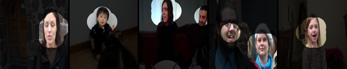}
&\includegraphics[width=0.325\linewidth]{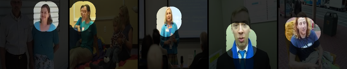}
&\includegraphics[width=0.325\linewidth]{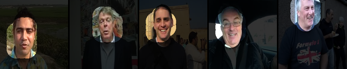}
\\
person&person
&person\\
\includegraphics[width=0.325\linewidth]{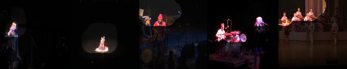}
&\includegraphics[width=0.325\linewidth]{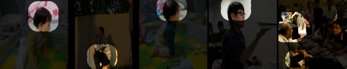}
&\includegraphics[width=0.325\linewidth]{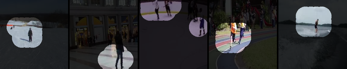}
\\
person&person&person\\
\includegraphics[width=0.325\linewidth]{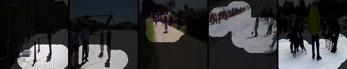}
&\includegraphics[width=0.325\linewidth]{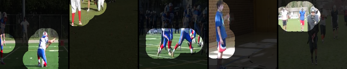}
&\includegraphics[width=0.325\linewidth]{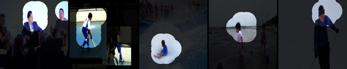}
\\
person&person&person\\
\includegraphics[width=0.325\linewidth]{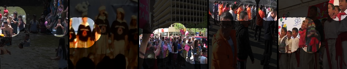}
&\includegraphics[width=0.325\linewidth]{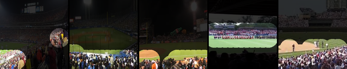}
&\includegraphics[width=0.325\linewidth]{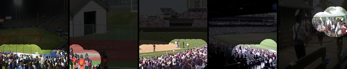}
\\
text&text&text\\
\includegraphics[width=0.325\linewidth]{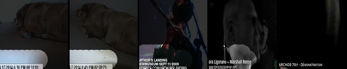}
&\includegraphics[width=0.325\linewidth]{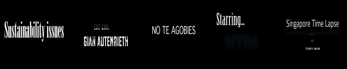}
&\includegraphics[width=0.325\linewidth]{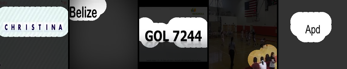}
\\
text
&baby&baby\\
\includegraphics[width=0.325\linewidth]{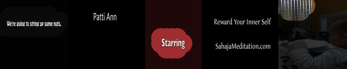}
&\includegraphics[width=0.325\linewidth]{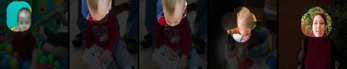}
&\includegraphics[width=0.325\linewidth]{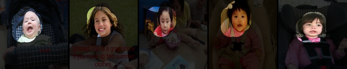}
\\
baby&baby
&baby\\
\includegraphics[width=0.325\linewidth]{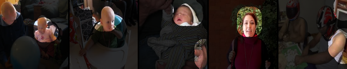}
&\includegraphics[width=0.325\linewidth]{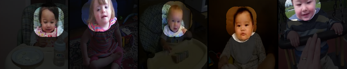}
&\includegraphics[width=0.325\linewidth]{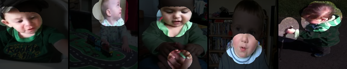}
\\
grass&grass&grass\\
\includegraphics[width=0.325\linewidth]{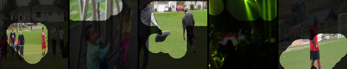}
&\includegraphics[width=0.325\linewidth]{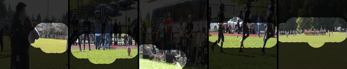}
&\includegraphics[width=0.325\linewidth]{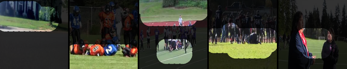}
\\
grass&waterfall&waterfall\\
\includegraphics[width=0.325\linewidth]{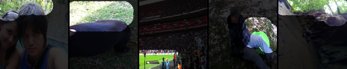}
&\includegraphics[width=0.325\linewidth]{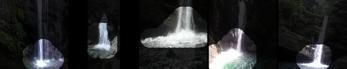}
&\includegraphics[width=0.325\linewidth]{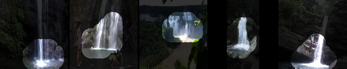}
\\
tree&tree
&tree\\
\includegraphics[width=0.325\linewidth]{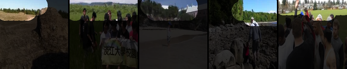}
&\includegraphics[width=0.325\linewidth]{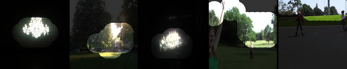}
&\includegraphics[width=0.325\linewidth]{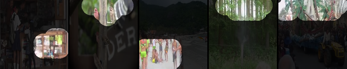}
\\
ceiling&ceiling&ceiling\\
\includegraphics[width=0.325\linewidth]{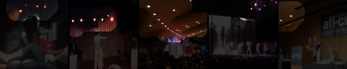}
&\includegraphics[width=0.325\linewidth]{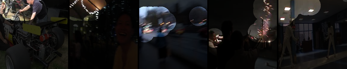}
&\includegraphics[width=0.325\linewidth]{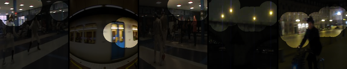}
\\
grandstand&grandstand
&car\\
\includegraphics[width=0.325\linewidth]{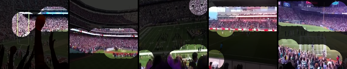}
&\includegraphics[width=0.325\linewidth]{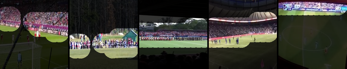}
&\includegraphics[width=0.325\linewidth]{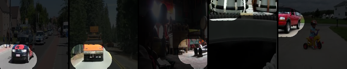}
\\
\midrule
\multicolumn{3}{c}{no object}\\
\includegraphics[width=0.325\linewidth]{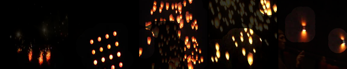}
&\includegraphics[width=0.325\linewidth]{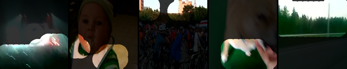}
&\includegraphics[width=0.325\linewidth]{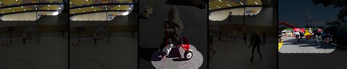}
\\
\multicolumn{3}{c}{no object}\\
\includegraphics[width=0.325\linewidth]{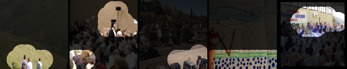}
&\includegraphics[width=0.325\linewidth]{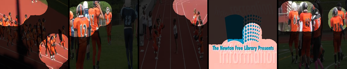}
&\includegraphics[width=0.325\linewidth]{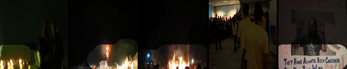}
\\
\end{tabular}
\caption{Top $5$ activations for units in our model ($39$ of \numobjunits from common classes). The last two rows show neurons that were not selective to an object class.}
\label{fig:allneuronvis}
\end{figure}

In \fig{fig:allneuronvis}, we provide visualizations of additional
object-selective neurons in our model. In \fig{fig:objdistrsupp} we
provide object-detector histograms for additional unsupervised methods
\cite{agrawal2015learning,wang2015unsupervised} (an extension of
\fig{fig:objdistr}).  We also show how the number of object-selective
units changes as a function of the threshold used to define whether
a unit is selective (\fig{fig:threshold-curve}).

\begin{figure}[t]
    \centering
  Training by sound (67 detectors)\\
  \includegraphics[width=\linewidth]{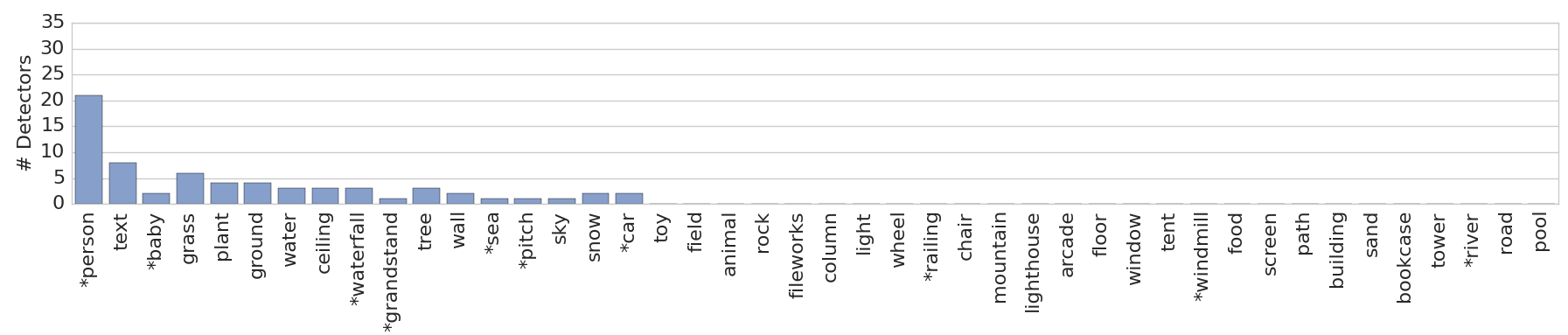} \\
  Training by labeled scenes~\cite{zhou2014places} (146 detectors)\\
  \includegraphics[width=\linewidth]{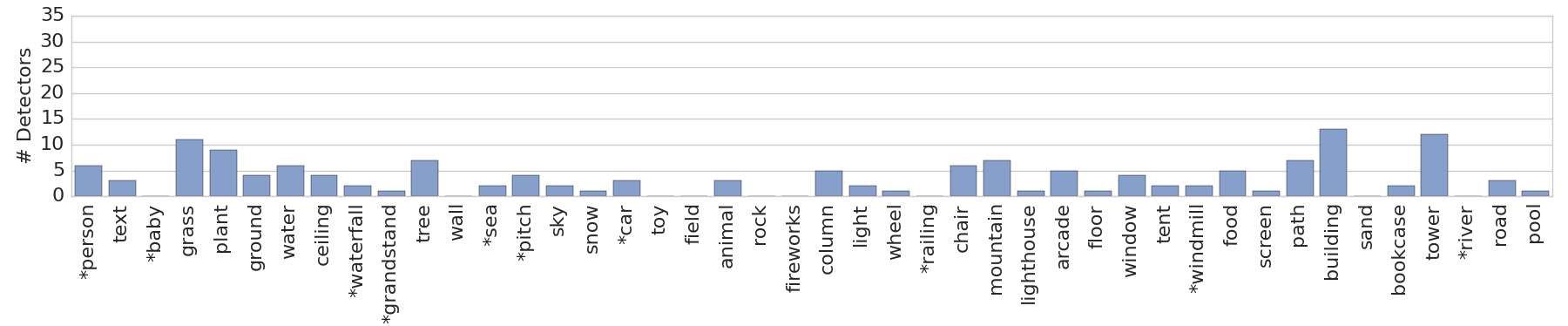} \\
  Training by visual tracking~\cite{wang2015unsupervised} (61 detectors)\\
  \includegraphics[width=\linewidth]{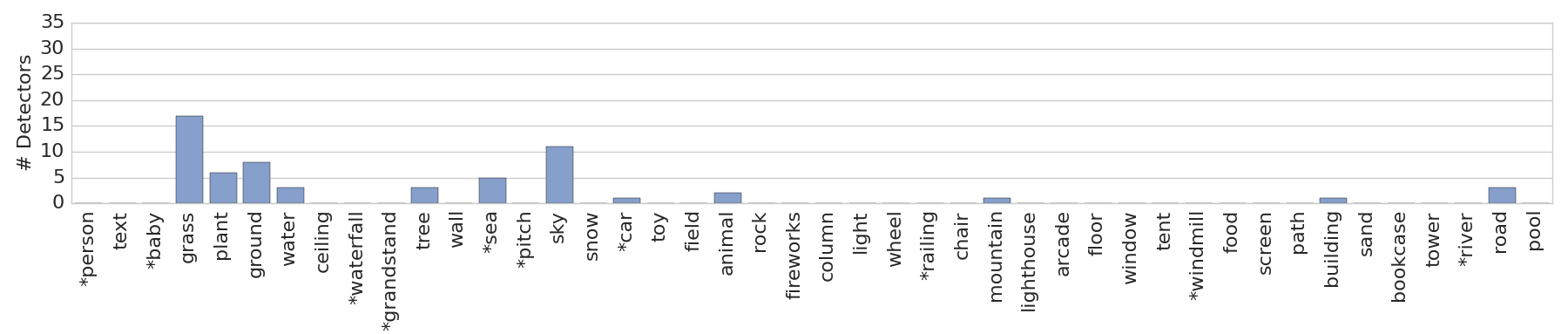}\\
  \caption{The number of object-selective per category, when
    evaluating the model on the SUN and ImageNet datasets
    (cf. \fig{fig:objdistr}, in which the models were evaluated on the
    Flickr video dataset).}
  \label{fig:objsun}
\end{figure}

To examine the effect of the dataset used to create the neuron
visualizations, we applied the same neuron visualization technique to
200,000 images sampled equally from the SUN and ImageNet datasets (as
in \cite{zhou2014object}). As expected, we found that the distribution
of objects was similar to that of the Flickr dataset
(\fig{fig:objsun}). Notably, there were fewer detectors in total
(\numobjsununits vs. \numobjunits), and there were some categories
(e.g., {\em baby}) that appeared relatively less often.  This may be
due to the differences in the underlying distribution of objects in
the datasets. For example, SUN focuses on scenes and contains more
objects labeled {\em tree}, {\em lamp}, and {\em window} than objects
labeled {\em person} \cite{zhou2014object}. We also computed a
detector histogram for the model of \cite{wang2015unsupervised},
finding that the total number of detectors was similar to the
sound-based model (\numobjtrackingsununits detectors), but that, as
before, the dominant categories were textural ``stuff'' (e.g., grass,
plants).

\begin{figure}[t]
\scriptsize\centering
\begin{tabular}{ccc}
person&car&ceiling\\
\includegraphics[width=0.325\linewidth]{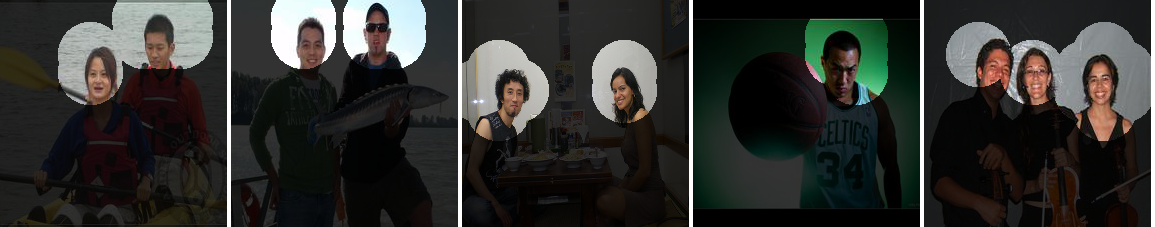}
&\includegraphics[width=0.325\linewidth]{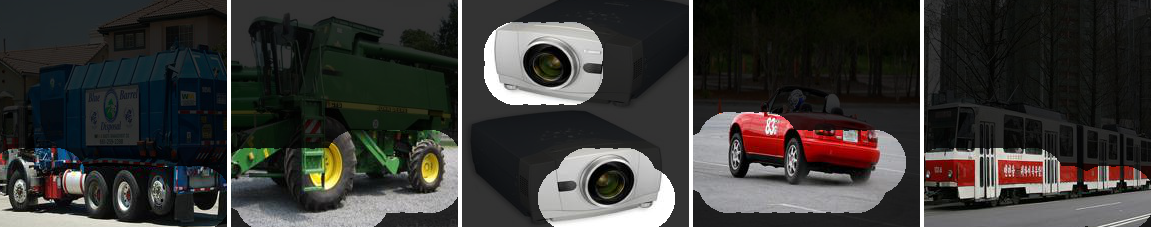}
&\includegraphics[width=0.325\linewidth]{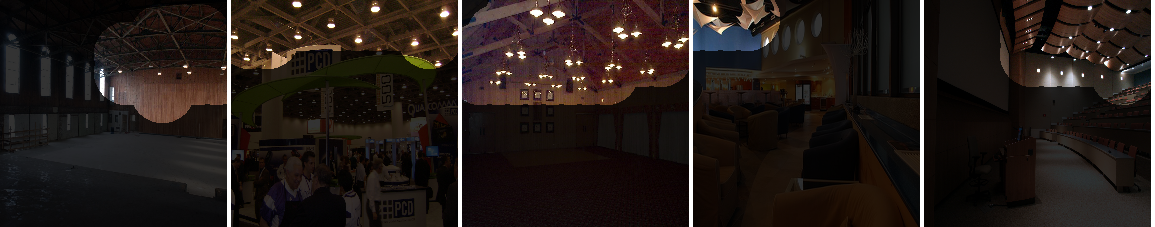}
\\
waterfall
&text&pitch\\
\includegraphics[width=0.325\linewidth]{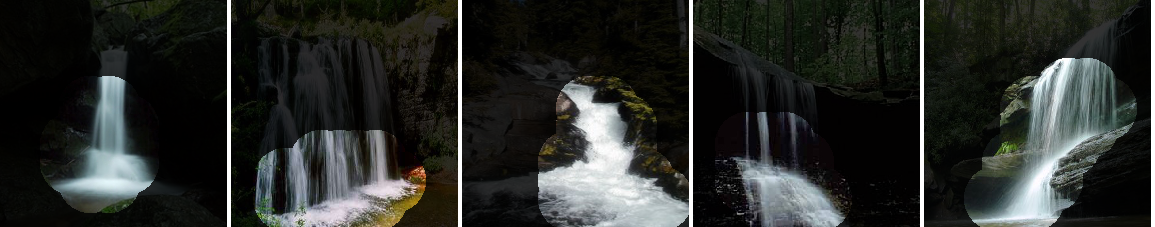}
&\includegraphics[width=0.325\linewidth]{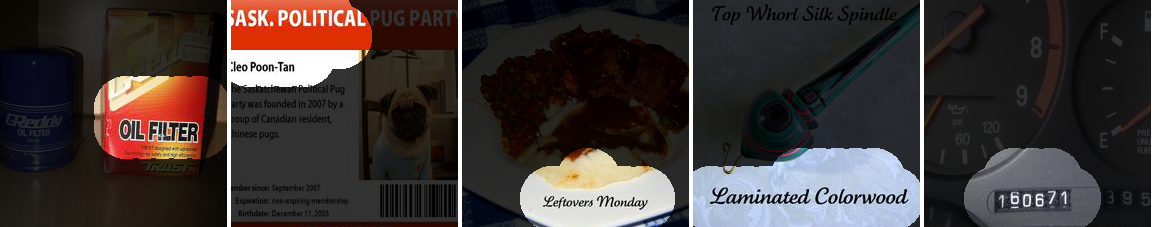}
&\includegraphics[width=0.325\linewidth]{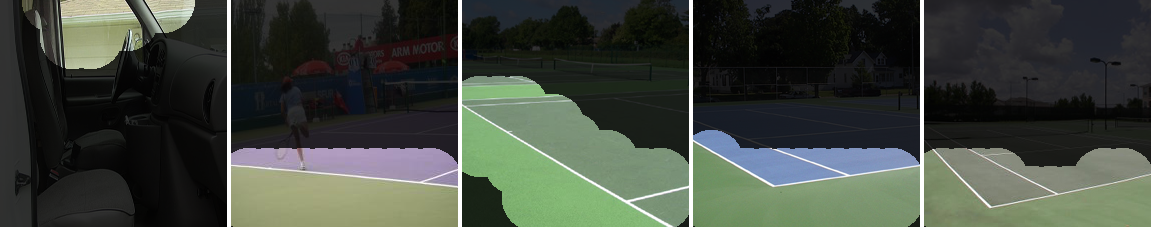}
\\
\end{tabular}
\caption{A selection of object-selective neurons, obtained by testing our model on the SUN and ImageNet datasets. We show the top 5 activations for each unit.} 
\end{figure}

\section{Sound textures}

We now describe, in more detail, how we computed sound textures from audio clips.  For this, we closely follow the work of McDermott and Simoncelli \cite{mcdermott2011sound}.

\vpar{Subband envelopes} To compute the cochleagram features $\{c_i\}$, we filter the input waveform $s$ with a bank of bandpass filters $\{f_i\}$.  
\begin{equation}
  c_i(t) = |(s \ast f_i) + j H(s \ast f_i)|,
\end{equation}
where $H$ is the Hilbert transform and $\ast$ denotes
cross-correlation. We then resample the signal to 400Hz and compress
it by raising each sample to the $0.3$ power (examples in
\fig{fig:soundtex}).

\vpar{Correlations} As described in Section 3, we compute the correlation between bands using a subset of the entries in the cochlear-channel correlation matrix.  Specifically, we include the correlation between channels $c_j$ and $c_k$ if $|j - k| \in \{1, 2, 3, 5\}$.  The result is a vector $\rho$ of correlation values.

\vpar{Modulation filters} We also include modulation filter responses. To get these, we compute each band's response to a filter bank $\{m_i\}$ of 10 bandpass filters whose center frequencies are spaced logarithmically from 0.5 to 200Hz:
\begin{equation}
  b_{ij} = \frac{1}{N}||c_i \ast m_j||^2,
\end{equation}
where $N$ is the length of the signal. 

\vpar{Marginal statistics}
We estimate marginal moments of the cochleagram features, computing the mean $\mu_i$ and standard deviation $\sigma_i$ of each channel.  We also estimate the loudness, $l$, of the sequence by taking the median of the energy at each timestep, i.e. $l = \mbox{median}(||c(t)||)$.

\vpar{Normalization} To account for global differences in gain, we normalize the cochleagram features by dividing by the loudness, $l$.  Following \cite{mcdermott2011sound}, we  normalize the modulation filter responses by the variance of the cochlear channel, computing $\tilde{b}_{ij} = \sqrt{\frac{b_{ij}}{\sigma_i^2}}$.
Similarly, we normalize the standard deviation of each cochlear channel, computing     $\tilde{\sigma}_{i} = \sqrt{\frac{\sigma_{i}^2}{\mu_i^2}}$.
From these normalized features, we construct a sound texture vector: $[\mu, \tilde{\sigma}, \rho, \tilde{b}, l]$

\end{document}